\documentclass[10pt,twocolumn,letterpaper]{article}
\usepackage[pagenumbers]{cvpr}

\makeatletter
\newcommand{\tightparagraphs}{\renewcommand\paragraph{\@startsection{paragraph}{4}{\z@}{0pt}{-1em}{\normalfont\normalsize\bfseries}}}
\makeatother
\makeatletter
\@ifpackageloaded{lineno}{%
  \IfHookEmptyTF{build/column/before}{}{\RemoveFromHook{build/column/before}[lineno]}%
  \AtBeginDocument{%
    \def\@LN@makecol{%
      \@LN@orig@makecol
      \setbox\@outputbox\vbox{\boxmaxdepth\@maxdepth
        \protected@write\@auxout{}{\string\def\string\@LN@column{\if@firstcolumn1\else2\fi}}%
        \box\@outputbox}}}%
}{}
\makeatother
\usepackage{multirow}
\usepackage{xspace}

\newcommand{\opt}[1]{\textcolor{cvprblue}{\boldsymbol{#1}}} 
\AtBeginDocument{\tightparagraphs} 

\usepackage{algorithm}
\usepackage{algpseudocode}

\definecolor{cvprblue}{rgb}{0.21,0.49,0.74}
\usepackage[pagebackref,breaklinks,colorlinks,allcolors=cvprblue]{hyperref}
\def\confName{CVPR}
\def\confYear{2027}

\title{OP2GS: Feature-Free Gaussian Splatting with Dual Opacity for 3D Instance Segmentation}
\author{%
Guiyu Liu$^{1}$ \quad Niklas Vaara$^{1}$ \quad Janne Mustaniemi$^{1}$ \quad Juho Kannala$^{1,2}$ \quad Janne Heikkil\"a$^{1}$\\
$^{1}$Center for Machine Vision and Signal Analysis, University of Oulu, Finland \quad $^{2}$Aalto University, Finland\vspace{-1.2em}
}

\begin{document}
\maketitle
\begin{abstract}
3D Gaussian Splatting (3DGS) provides an explicit and efficient scene representation, but its primitives do not naturally expose object-level identity.
Existing methods either learn contrastive per-Gaussian features that need post-hoc clustering, or supervise identity encodings with view-consistent 2D masks.
We propose OP2GS, a lightweight object-aware Gaussian representation that augments each primitive with an explicit instance label and a dedicated instance opacity $\sigma^{*}$, which serves as an instance occupancy field for object-mask rendering.
This dual-opacity formulation decouples visual appearance from instance occupancy, so that $\sigma^{*}$ can be learned and the instance labels can be corrected without touching the appearance parameters $\sigma$ and $c$.
We learn $\sigma^{*}$ with an instance occupancy loss that follows the standard transmittance-based visibility of 3DGS.
A discrete regrouping step then corrects the instance labels using multi-view mask consensus.
Since no high-dimensional features are required, OP2GS can be plugged into existing 3DGS training pipelines and applied unchanged to frozen 3DGS assets.
On the ScanNet++ validation set with the official class-agnostic protocol, OP2GS improves over OpenSplat3D, the strongest comparable baseline, with both SAM and EntitySeg masks; it is also best on Messy Rooms and, under the Gaga protocol, on Replica, while storing only one scalar and one integer per Gaussian. Code will be released.
\end{abstract}

\section{Introduction}
\label{sec:intro}

\begin{figure}[t]
  \centering
  \includegraphics[width=0.92\linewidth]{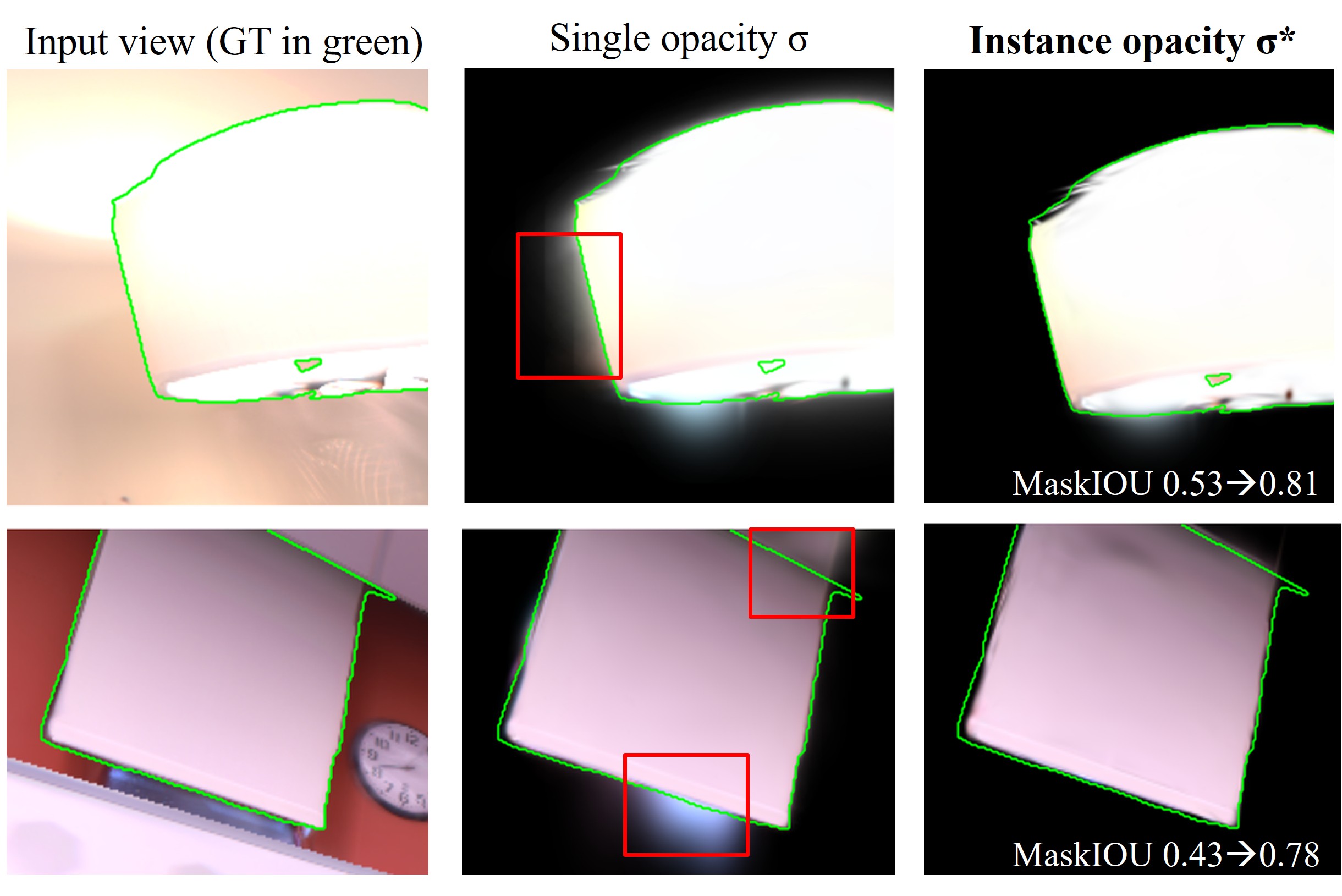}
  \caption{\textbf{Why a second opacity.} With the single opacity $\sigma$, the mask of two Replica objects leaks into boundary Gaussians and floaters that the image needs (red boxes); the instance opacity $\sigma^{*}$ removes them while $\sigma$ and the image stay unchanged. PSNR of the extracted object on the test views: 17.0\,$\to$\,22.0\,dB (top), 16.9\,$\to$\,22.5\,dB (bottom).}
  \label{fig:teaser}
\end{figure}

3D Gaussian Splatting (3DGS)~\cite{kerbl20233dgs} has emerged as an efficient explicit representation for real-world scenes, but its original primitive is defined only by geometry, color, and opacity, without any notion of object identity.
Zero-shot 3D instance segmentation of a 3DGS scene, \ie, lifting the class-agnostic masks of a 2D foundation model such as SAM~\cite{sam_Kirillov_2023_ICCV} into 3D objects without 3D supervision, is therefore a basic step for downstream tasks such as object-level editing.
Existing methods give the Gaussians an identity in one of two ways: by training a per-Gaussian feature, or by lifting 2D masks onto the Gaussians without training.

Feature-based methods attach a feature vector to each Gaussian, render it with the same weights as the color, and supervise the rendered feature with 2D masks.
Contrastive methods~\cite{wu2024opengaussian,piekenbrinck2025opensplat3d,zhu2025unifiedlift,li2025instancegaussian,clickgaussian2024} recover instances afterwards by clustering the features.
Identity-encoding methods~\cite{ye2023gaussiangrouping,zhu2025objectgs} decode the feature with a classifier instead; they need masks that are already consistent across views, obtained from a video tracker such as DEVA~\cite{cheng2023deva} or SAM2~\cite{ravi2025sam2}.
In both cases the features increase memory consumption at render time, and high-dimensional features~\cite{qin2024langsplat,ye2023gaussiangrouping} additionally need a decoder network to be read out.
More importantly, the features are trained together with the reconstruction; on a frozen, already reconstructed 3DGS asset they can only be trained afterwards, which needs many more iterations to fit the given labels and, with SAM masks, costs 1.3 AP (\cref{sec:exp_capability}).

Lifting-based methods~\cite{lyu2024gaga,shen2024flashsplat,cheng2024occam,jun2025drsplat,bang2026lightsplat} avoid training altogether and assign each Gaussian the label of the 2D mask that receives most of its rendering weight.
Because Gaussians are semi-transparent and alpha-composited, a 2D mask does not map cleanly onto a set of Gaussians: Gaussians on object boundaries and floaters project into a mask with substantial weight and receive its label although they are not part of the object.
The label is a hard integer that receives no gradient, so such errors cannot be corrected by optimization.

Both families share one cause: the single opacity $\sigma$ of 3DGS serves two roles at once.
It decides how much a Gaussian contributes to the image, and, because features are rendered and masks are lifted with the rendering weights of $\sigma$, it also decides how much the Gaussian contributes to the mask of its object.
The two roles do not need to agree: a floater produced by the reconstruction should not appear in any object mask (\cref{fig:teaser}), but with a single opacity the only way to remove it from the mask is to lower $\sigma$, which also removes it from the image (\cref{sec:exp_ablation}).
We therefore propose to give each Gaussian a second opacity for this role.
Besides $\sigma$ for appearance, each Gaussian carries an integer instance label, which says \emph{which} object it belongs to, and an instance opacity $\sigma^{*}$, which says \emph{how much} it contributes to the mask of that object.
We call the resulting representation OP2GS.
Because $\sigma^{*}$ is an opacity, it is defined on a fixed $[0,1]$ scale.

We learn $\sigma^{*}$ with an \emph{instance occupancy loss} that renders the mask of an object with $\sigma^{*}$ but with the transmittance of the original opacity $\sigma$, so that occlusion is inherited from the appearance while the gradient flows only into $\sigma^{*}$.
The labels cannot be learned by gradients.
We propose a training-free 3D association that clusters the per-frame masks into scene-level objects by multi-view co-visibility rather than by a video tracker, and lifts the clusters onto the Gaussians by voting.
The lifted labels contain errors, and lowering $\sigma^{*}$ can only remove a Gaussian from its object, not move it to the right one.
We therefore propose a discrete regrouping step that partitions the Gaussians into single-label superpoints and merges them under multi-view mask consensus and a prior from the lifted labels (\cref{fig:pipeline}); afterwards, $\sigma^{*}$ is retrained.

Our main contributions are:
\begin{itemize}
    \item \textbf{Dual-opacity representation.}
    We augment each Gaussian with an integer instance label and an instance opacity $\sigma^{*}$, which decouples object-mask rendering from visual reconstruction. Because of this decoupling, segmentation is learned without touching the appearance parameters $\sigma$ and $c$ (PSNR within 0.04\,dB of an RGB-only 3DGS, \cref{tab:dual}), which is what allows the same pipeline to run unchanged on existing, even frozen, 3DGS assets (\cref{tab:frozen}).
    \item \textbf{Segmentation pipeline for feature-free Gaussians.}
    On top of this representation, we propose a training-free mask association, instance occupancy learning with a loss that reuses the transmittance of $\sigma$, and a discrete label regrouping based on multi-view mask consensus; no post-hoc clustering and no decoding network are needed.
    \item \textbf{Competitive results on four benchmarks.}
    On the ScanNet++ validation set~\cite{yeshwanth2023scannetpp}, OP2GS reaches 20.2 AP with SAM masks against 19.2 for the strongest comparable baseline~\cite{piekenbrinck2025opensplat3d}, and 23.1 against 18.3 AP with EntitySeg masks; it also reaches the highest mIoU on ScanNet~\cite{dai2017scannet} and, under the Gaga protocol, on Replica~\cite{straub2019replica}, the highest mean PQ$^\text{scene}$ on Messy Rooms~\cite{bhalgat2023contrastivelift}, and loses at most 0.4 AP on frozen 3DGS assets.
\end{itemize}

\section{Related Work}
\label{sec:related}

We review the three lines of work that our pipeline builds on: lifting 2D masks to 3D, which is independent of the representation; feature fields in radiance fields and 3DGS; and training-free segmentation and object-level representations on 3DGS.

\paragraph{Lifting 2D masks to 3D.}
On point clouds and meshes, zero-shot 3D instance segmentation merges per-frame SAM masks into 3D instances without 3D training: SAM3D~\cite{yang2023sam3d} merges masks bottom-up along the frame sequence, SAI3D~\cite{yin2024sai3d} grows geometric primitives under SAM guidance, MaskClustering~\cite{yan2024maskclustering} clusters a mask graph by view consensus, SAM2Object~\cite{zhao2025sam2object} merges superpoints with mask affinities, and Open3DIS~\cite{nguyen2024open3dis} adds open-vocabulary features; Segment3D~\cite{huang2024segment3d} goes one step further and trains a 3D model on the automatically generated masks.
The central problem in all of them is the same: per-frame masks carry no common identity across views.
Alternatively, consistent identities come from a video tracker such as DEVA~\cite{cheng2023deva} or SAM2~\cite{ravi2025sam2}, on which Gaussian Grouping~\cite{ye2023gaussiangrouping} and ObjectGS~\cite{zhu2025objectgs} rely; tracking assumes a continuous trajectory and can lose consistency under the large viewpoint changes between frames of real scans.
Panoptic Lifting~\cite{siddiqui2023panoptic} instead re-matches ids by linear assignment at every optimization step.
Our association stage transfers the view-consensus criteria to 3D Gaussians, where the rendering weights replace point visibility, and treats the result only as an initialization that the subsequent occupancy learning and regrouping correct.

\paragraph{Feature fields in radiance fields and 3DGS.}
Semantic-NeRF~\cite{zhi2021semanticnerf} and Panoptic Lifting~\cite{siddiqui2023panoptic} show that noisy 2D predictions can be fused into a neural field to obtain view-consistent semantic and instance maps, DFF~\cite{kobayashi2022dff} and LERF~\cite{kerr2023lerf} distill 2D features into the field for editing and open-vocabulary queries, and Contrastive Lift~\cite{bhalgat2023contrastivelift} learns an instance embedding field with a slow--fast contrastive loss, introducing the Messy Rooms benchmark.

With 3DGS~\cite{kerbl20233dgs}, the same idea is realized by rendering a feature vector per Gaussian alongside its color, in three forms.
Language features~\cite{qin2024langsplat,shi2024legaussians,zhou2024feature3dgs,wu2024opengaussian,jang2025ilgs} distill CLIP and need a scene-wise autoencoder or quantization to keep storage tractable; SceneSplat~\cite{li2025scenesplat} predicts them feed-forward for unseen scenes.
Contrastive instance features are trained so that Gaussians of one object cluster together: OpenSplat3D~\cite{piekenbrinck2025opensplat3d} clusters 8-dimensional features with HDBSCAN~\cite{campello2013hbscan}, InstanceGaussian~\cite{li2025instancegaussian} jointly learns appearance and 6-dimensional instance features, Unified-Lift~\cite{zhu2025unifiedlift} and UniC-Lift~\cite{dhiman2026uniclift} learn them end-to-end with a codebook, Click-Gaussian~\cite{clickgaussian2024}, OmniSeg3D~\cite{ying2024omniseg3d}, and EgoLifter~\cite{gu2024egolifter} learn two-level, hierarchical, or transient-aware fields, and Trace3D~\cite{shen2025trace3d} splits and deletes ambiguous Gaussians before learning such features, thereby editing the reconstruction itself.
Identity encodings replace the embedding by a decodable code: Gaussian Grouping~\cite{ye2023gaussiangrouping} renders a 16-dimensional encoding decoded by a classifier, and ObjectGS~\cite{zhu2025objectgs} renders one-hot labels whose length equals the number of objects.
OP2GS renders no feature: each Gaussian stores an integer label and a scalar instance opacity, and instances are read out without clustering or a decoder.

\paragraph{Training-free segmentation and object-level 3DGS.}
A third line keeps the trained 3DGS fixed and assigns labels or features to its Gaussians without optimization: Gaga~\cite{lyu2024gaga} associates masks through a 3D-aware memory bank, FlashSplat~\cite{shen2024flashsplat} solves the per-Gaussian assignment by linear programming, Occam's LGS~\cite{cheng2024occam} and DR-Splat~\cite{jun2025drsplat} lift language features through the rendering weights, LUDVIG~\cite{marrie2025ludvig} and VALA~\cite{wang2025vala} aggregate 2D features over views with visibility weighting, and LightSplat~\cite{bang2026lightsplat} stores a compact cluster index per Gaussian.
Object-level representations use such assignments for editing~\cite{ye2023gaussiangrouping,qiu2024featuresplatting,cen2025saga}, and Segment-then-Splat~\cite{lu2025segmentthensplat} partitions the Gaussians into object-specific sets before reconstruction; OP2GS keeps a single reconstruction and learns and corrects membership after the scene is built, including on frozen assets.
In the training-free methods, the only quantity tying a Gaussian to a mask is the rendering weight of its appearance opacity (\cref{sec:intro}).
OP2GS also starts from a rendering-weight-based assignment (\cref{sec:assoc}) but adds a learnable instance opacity that suppresses wrongly assigned Gaussians without touching the appearance parameters, and a discrete regrouping (\cref{sec:regroup}) whose edge weights combine multi-view mask agreement with the lifted labels.

\section{Method}
\label{sec:method}
\begingroup\tightparagraphs

\begin{figure*}[t]
  \centering
  \includegraphics[width=\linewidth]{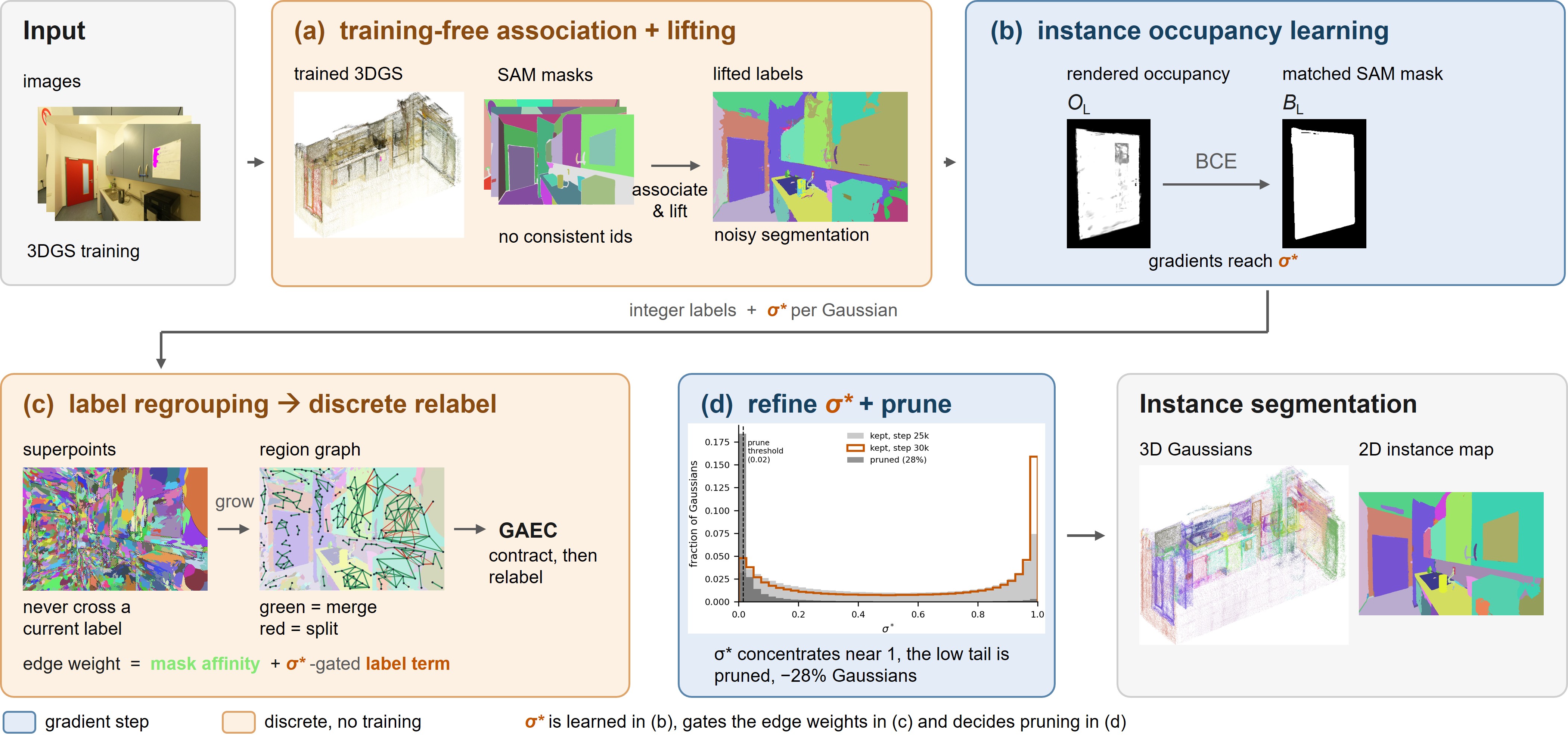}
  \caption{\textbf{Overview of OP2GS.} Each Gaussian carries an integer instance label $l$ and an instance opacity $\sigma^{*}$ in addition to the usual 3DGS parameters. (a)~A training-free association lifts 2D masks into initial, noisy labels. (b)~The instance occupancy loss trains $\sigma^{*}$ by matching the rendered object mask to the 2D mask. (c)~Regrouping corrects the labels by region growing on superpoints and graph partitioning with GAEC. (d)~A final round refines $\sigma^{*}$ and optionally prunes; the threshold in the inset (0.02 in this scene) is the 10\% quantile of $\sigma^{*}$, and together with Gaussians that are never rendered 28\% of the Gaussians are removed. Gradients never touch the appearance opacity $\sigma$ or the colors, so the same pipeline also runs on a frozen 3DGS asset.}
  \label{fig:pipeline}
\end{figure*}

Given posed images $\{I_t\}$ of a static scene and class-agnostic SAM~\cite{sam_Kirillov_2023_ICCV} masks for each image, our goal is to assign every Gaussian to an object instance, which also yields view-consistent 2D instance maps.
We first define the dual-opacity representation (\cref{sec:primitive}); the rest of the pipeline only assigns values to its two new parameters: a training-free association produces initial labels (\cref{sec:assoc}), the instance occupancy loss trains $\sigma^{*}$ (\cref{sec:occupancy}), a discrete regrouping corrects the labels (\cref{sec:regroup}), and a final round retrains $\sigma^{*}$ and prunes Gaussians with low $\sigma^{*}$ (\cref{sec:refine}).

\subsection{OP2GS Representation}
\label{sec:primitive}

A 3DGS~\cite{kerbl20233dgs} scene is a set $\mathcal{G}_n$ of $n$ Gaussians with center $\mu$, color $c$, opacity $\sigma$, rotation $R$ and scale $S$, rasterized by a differentiable renderer $\mathcal{D}$ and trained with a photometric loss $\mathcal{L}_\text{rgb}$.
OP2GS extends each Gaussian with an integer instance label $l$ and an instance opacity $\sigma^{*} \in [0,1]$.
Instead of a high-dimensional feature that encodes identity~\cite{wu2024opengaussian,piekenbrinck2025opensplat3d,ye2023gaussiangrouping}, a Gaussian thus stores one scalar and one integer, and the same renderer additionally produces an instance occupancy map $O_L$ for any object $L$, supervised by a binary mask $B_L$; optimizable parameters are shown in bold:
\begin{equation}
  \begin{aligned}
    \mathcal{G}(\opt{\mu}, \opt{c}, \opt{\sigma}, \opt{R}, \opt{S}, \opt{\sigma^{*}}, l)_n
      &\xrightarrow{\mathcal{D}} O_L,\, I_\text{ren} \\
      &\longrightarrow \mathcal{L}_\text{occ},\, \mathcal{L}_\text{rgb} \longleftarrow B_L,\, I_\text{gt}.
  \end{aligned}
  \label{eq:overview}
\end{equation}
The label $l$ in \cref{eq:overview} receives no gradient; it is set by the association (\cref{sec:assoc}) and rewritten by the regrouping (\cref{sec:regroup}).
We call the extraction of instances from the representation the \emph{readout}: a 3D instance is the set of Gaussians sharing a label with $\sigma^{*}_i > \tau$ ($\tau = 0.1$), and its 2D mask in a view is read out from the rendered occupancy $O_L$ (\cref{sec:occupancy}), either as the label with the largest $O_L(v)$ at each pixel (\emph{sum readout}) or as the label of the Gaussian with the largest $\alpha^{*}_i T_i$ along the ray (\emph{max readout}); no clustering is involved.
All hyperparameters introduced below, including $\tau$, are shared by all datasets and listed in the supplementary.

\subsection{Training-Free Association and Lifting}
\label{sec:assoc}

We first train the 3DGS model with $\mathcal{L}_\text{rgb}$ only, for the first 20k of the 30k iterations.
To label its Gaussians, we lift the SAM masks onto them.
SAM masks, however, are produced independently per view: the same object has a different id in every view and is often split into parts.
We therefore first associate the masks across views and then lift the aligned ids onto the Gaussians.

\paragraph{Association.}
We adopt the idea of MaskClustering~\cite{yan2024maskclustering} to merge per-view masks by multi-view consensus, but define the link between a mask and the scene through the rendering weights of the Gaussians instead of point membership.
The association only produces initial labels.
(1)~\emph{Linking masks to Gaussians.}
For every mask $m$ in view $t$, we render the view with the trained model and keep the Gaussians whose rendering weight $\alpha_i T_i$ falls almost entirely (at least 80\%) inside $m$.
These Gaussians, the \emph{core} of $m$, are what the mask covers in 3D; boundary Gaussians and floaters, whose weight is spread over several masks, are left out.
(2)~\emph{Projection into other views.}
Each mask starts as its own group and is projected into every other view through its core Gaussians: the group is \emph{visible} in a view if at least half of its core receives rendering weight there, and it is \emph{attributed} to the mask of that view that receives most of this weight.
(3)~\emph{Merging by co-visibility consensus.}
The consensus of two groups is the fraction of views in which both are visible and attributed to the same mask.
A view in which one mask merely contains the other does not count, so that small objects are not absorbed into room-scale masks.
Groups are merged iteratively, with the consensus threshold decreasing from 0.9 to 0.7 so that confident merges happen first; groups seen in fewer than three views are discarded.
Every remaining group receives a global id, and the pixels of each mask are rewritten with the id of its group, which gives view-consistent masks.
The SAM mask extraction and the thresholds of the association are detailed in the supplementary.

\paragraph{Lifting.}
Every Gaussian $i$ then receives the global id with the largest \emph{share} of its rendering weight, $l_i = \arg\max_k S_{ik}/A_{ik}$, where $S_{ik}$ is the rendering weight $\alpha_i T_i$ of $i$ summed over all pixels of id $k$ in all views and $A_{ik}$ is its total rendering weight in the views in which $k$ appears; ids that receive weight from $i$ in fewer than three views are not considered. Voting by share rather than by total weight prevents large or frequently seen objects from winning by size.
Gaussians without any vote take the label of their nearest labeled neighbor.
Finally, the instance opacity is initialized as $\sigma^{*}_i = \sigma_i$, so that the instance branch initially reproduces the hard lifting result, including its errors, which the next two stages are designed to reduce.

\subsection{Instance Occupancy Learning}
\label{sec:occupancy}

This stage learns the instance opacity $\sigma^{*}_i$ of every Gaussian, \ie, its contribution to the mask of its labeled object, so that it suppresses the contribution of wrongly labeled Gaussians.

\paragraph{Rendering the instance occupancy.}
For a pixel $v$, 3DGS sorts the overlapping Gaussians front to back, indexed $i=1,2,\dots$, and composites their colors,
\begin{equation}
  C(v) = \sum_{i} c_i\, \alpha_i T_i, \qquad \alpha_i = \sigma_i\, G_i(v),
  \label{eq:color}
\end{equation}
where $G_i(v)$ is the projected 2D Gaussian evaluated at $v$ and $T_i = \prod_{j<i}(1-\alpha_j)$ the transmittance.
The instance occupancy of object $L$ at $v$ is composited along the same ray with the instance opacity in place of $\sigma$, but with the transmittance of \cref{eq:color}:
\begin{equation}
  O_L(v) = \min\Bigl(1,\, \sum_{i:\, l_i = L} \alpha^*_i\, T_i\Bigr), \qquad \alpha^*_i = \sigma^{*}_i\, G_i(v).
  \label{eq:occupancy}
\end{equation}
This formulation has three properties.
First, $T_i$ is computed from the appearance opacity $\sigma$, so occlusion is inherited from the appearance: a Gaussian hidden behind another object in the image is also hidden in the mask.
Second, the sum runs only over Gaussians with label $L$ and weights them by $\sigma^{*}_i$ rather than $\sigma_i$, so $O_L$ is the mask of one object, and the contribution of a Gaussian to this mask is independent of its contribution to the image.
Third, $G_i$ and $T_i$ are shared with the color rendering, so the only new quantity is $\sigma^{*}_i$.
Since $\sigma^{*}_i$ is not bounded by $\sigma_i$, the sum can exceed one and is clamped so that $O_L$ can be supervised with binary masks.

\paragraph{Supervision.}
The view-consistent masks of \cref{sec:assoc} were needed only to make the lifted labels consistent; for supervision we use the raw per-view SAM masks.
At each training step, we render the instance map of the current view with the max readout and match its label regions one-to-one to the SAM masks of that view: the Hungarian algorithm on mask IoU pairs every label with at most one mask and every mask with at most one label, and pairs with IoU $< 0.3$ are discarded.
From the matched labels we select a set $\mathcal{M}$ of $M$ labels, giving priority to labels supervised least often so far, and apply the \emph{instance occupancy loss}
\begin{equation}
  \mathcal{L}_\text{occ} = \frac{1}{M}\sum_{L \in \mathcal{M}} \frac{1}{|V|}\sum_{v \in V} \mathrm{BCE}\bigl(O_L(v),\, B_L(v)\bigr),
  \label{eq:occloss}
\end{equation}
where $B_L$ is the binary SAM mask matched to label $L$ and $V$ is the set of pixels of the view.
Gaussians of $L$ that are visible inside $B_L$ are pushed toward $\sigma^{*} = 1$; those visible outside $B_L$, including at pixels of other objects, are treated as negatives and pushed toward $\sigma^{*} = 0$.
The total loss is $\mathcal{L} = \mathcal{L}_\text{rgb} + \lambda_\text{occ} \mathcal{L}_\text{occ}$ with $\lambda_\text{occ} = 0.1$.

\paragraph{Gradient paths.}
The two losses update different parameters.
$\mathcal{L}_\text{rgb}$ updates $\sigma_i$, $c_i$ and the geometry $\mu_i, R_i, S_i$ through $G_i(v)$, as in 3DGS.
For $\mathcal{L}_\text{occ}$ we stop the gradient through $T_i$, so it never reaches $\sigma$ or the colors; for a Gaussian $i$ with label $L = l_i \in \mathcal{M}$, the gradient with respect to its instance opacity is
\begin{equation}
  \frac{\partial \mathcal{L}_\text{occ}}{\partial \sigma^{*}_i} = \sum_{v \in V} \frac{\partial \mathcal{L}_\text{occ}}{\partial O_L(v)}\, G_i(v)\, T_i\, \mathbf{1}_v,
  \label{eq:grad}
\end{equation}
where $\mathbf{1}_v = 1$ if the sum of \cref{eq:occupancy} at $v$ is below one and $0$ where it is clamped; an occluded Gaussian ($T_i \approx 0$) receives no gradient.
In joint training, $\mathcal{L}_\text{occ}$ also updates the geometry through $G_i(v)$, so both losses shape $\mu, R, S$ while appearance ($\sigma, c$) and occupancy ($\sigma^{*}$) are optimized independently; the effect on rendering is within 0.04\,dB PSNR (\cref{tab:dual}).
After this stage, $\sigma^{*}$ is a per-Gaussian confidence in its own label.

\subsection{Label Regrouping}
\label{sec:regroup}

Occupancy learning can only lower the contribution of a Gaussian to the object it is labeled with.
Because the labels are not trainable, it cannot merge two labels that cover one object or split one that covers two, the typical errors of lifting; regrouping rewrites the labels in three steps.
(1)~To make splitting possible, the Gaussians are over-segmented into \emph{superpoints}~\cite{felzenszwalb2004efficient}, spatially connected groups with one lifted label each, so that a label becomes a set of superpoints that can be regrouped independently.
(2)~Superpoints are merged into \emph{regions} by region growing on their \emph{mask affinity} $a_{pq} \in [0,1]$: in a view, $f_p$ is the vector of fractions of the visible Gaussians of $p$ that fall into each SAM mask, and $a_{pq}$ is the cosine similarity of $f_p$ and $f_q$, averaged over co-visible views; the lifted labels are ignored here, so superpoints of two labels that cover one object are joined, and superpoints of one label that cover two objects stay apart.
(3)~Regions are merged by correlation clustering on a signed graph~\cite{keuper2015gaec}, which adds the lifted labels back as a prior: two regions $p$ and $q$ are connected if any of their Gaussians are neighbors in the $k$-NN graph, with edge weight
\begin{equation}
  w_{pq} = 2\,(a_{pq} - 0.5) \;+\; \lambda_l\, \sqrt{\bar\sigma^*_p\, \bar\sigma^*_q}\; s_{pq},
  \label{eq:edge}
\end{equation}
where $s_{pq}{=}{+}1$ if $p$ and $q$ carry the same lifted label and $-1$ otherwise, $\bar\sigma^*_p$ is the mean instance opacity of $p$, and $\lambda_l = 0.5$ weights the label prior.
The first term lets mask evidence merge ($a_{pq} > 0.5$) or separate ($a_{pq} < 0.5$) regions; the second is a prior toward the lifted grouping, weighted by the mean $\sigma^{*}$ of the two regions, which is low for labels that mix objects.
Greedy additive edge contraction (GAEC)~\cite{keuper2015gaec} contracts the most positive edge until none remains and needs no target number of instances.
Each resulting group becomes a new label, and Gaussians whose label changed restart with $\sigma^{*} = \sigma$.

\subsection{Refinement}
\label{sec:refine}

After regrouping, we continue training $\sigma^{*}$ with \cref{eq:occloss} in a second stage, with the same loss but different targets.
We render the instance map as before, but choose \emph{many-to-one} targets rather than one-to-one Hungarian matching: the target $B_L$ of a selected label $L$ is the union of all SAM masks that fall mostly inside its rendered region, so that a label may be matched to several masks.
Gaussians that receive no rendering weight in any training view, or whose $\sigma^{*}$ falls in the lowest 10\%, are optionally removed during this stage (three times, after 1k, 2k, and 3k of its 5k iterations); this does not change the filtered readout (20.2 AP with and without pruning, \cref{tab:readout}).
\endgroup

\section{Experiments}
\label{sec:exp}
\begingroup\tightparagraphs
\begin{table}[t]
\centering
\caption{\textbf{ScanNet++}, 50 validation scenes, official evaluation on mesh vertices with official ignore masks, all scenes pooled (\%). Class-aware results assign one of 84 instance classes with MasQCLIP~\cite{xu2023masqclip}. $^\dagger$our re-run with the official code, same frames, masks and evaluation as OP2GS; ``pub.'': the published OpenSplat3D numbers without post-processing. SAM3D and Segment3D numbers are taken from OpenSplat3D~\cite{piekenbrinck2025opensplat3d}. SGIFormer is a fully supervised point-cloud method (``sup.''), listed for reference only.}
\label{tab:scannetpp}
\setlength{\tabcolsep}{1.8pt}
\footnotesize
\begin{tabular}{@{}llcccccc@{}}
\toprule
 & & \multicolumn{3}{c}{Class-agnostic} & \multicolumn{3}{c}{Class-aware} \\
\cmidrule(lr){3-5}\cmidrule(l){6-8}
Method & Masks & AP & AP$_{50}$ & AP$_{25}$ & AP & AP$_{50}$ & AP$_{25}$ \\
\midrule
\textcolor{gray}{SGIFormer~\cite{yao2025sgiformer} (sup.)} & \textcolor{gray}{--} & \textcolor{gray}{--} & \textcolor{gray}{--} & \textcolor{gray}{--} & \textcolor{gray}{23.9} & \textcolor{gray}{37.5} & \textcolor{gray}{46.6} \\
\midrule
SAM3D~\cite{yang2023sam3d} & SAM & 3.9 & 9.3 & 22.1 & -- & -- & -- \\
Segment3D~\cite{huang2024segment3d} & -- & 13.0 & 23.8 & 38.3 & -- & -- & -- \\
\midrule
OpenSplat3D~\cite{piekenbrinck2025opensplat3d} (pub.) & SAM & 19.2 & 37.3 & 56.2 & -- & -- & -- \\
OpenSplat3D$^\dagger$ & SAM & 19.2 & \textbf{37.4} & \textbf{56.9} & 13.6 & \textbf{28.2} & 39.2 \\
OP2GS & SAM & \textbf{20.2} & 37.1 & 54.7 & \textbf{14.1} & 27.6 & \textbf{40.7} \\
\midrule
OpenSplat3D$^\dagger$ & EntitySeg & 18.3 & 36.3 & 55.2 & 14.1 & 30.0 & 40.0 \\
OP2GS & EntitySeg & \textbf{23.1} & \textbf{40.9} & \textbf{56.8} & \textbf{17.6} & \textbf{33.0} & \textbf{42.1} \\
\bottomrule
\end{tabular}
\end{table}

\begin{table}[t]
\centering
\caption{\textbf{ScanNet}, 10 scenes (\%). Baselines report only mAcc@0.25; Gaussian Grouping and OpenGaussian numbers are those reported by InstanceGaussian~\cite{li2025instancegaussian}. $^\dagger$our re-evaluation of the released InstanceGaussian predictions.}
\label{tab:scannet}
\setlength{\tabcolsep}{2pt}
\begin{tabular}{@{}llccc@{}}
\toprule
Method & Masks & mIoU & \begin{tabular}[b]{@{}c@{}}mAcc\\@.5\end{tabular} & \begin{tabular}[b]{@{}c@{}}mAcc\\@.25\end{tabular} \\
\midrule
Gaussian Grouping~\cite{ye2023gaussiangrouping} & SAM+DEVA & 22.55 & -- & 30.54 \\
OpenGaussian~\cite{wu2024opengaussian} & SAM & 27.32 & -- & 52.44 \\
Ilov3Splat~\cite{ilov3splat2026} & SAM & 32.08 & -- & 59.29 \\
InstanceGaussian~\cite{li2025instancegaussian} & SAM & 50.27 & -- & 80.22 \\
InstanceGaussian$^\dagger$ & SAM & 50.30 & \textbf{52.58} & 80.25 \\
OP2GS & EntitySeg & 51.52 & 51.96 & 83.77 \\
OP2GS & SAM & \textbf{51.90} & 49.57 & \textbf{87.90} \\
\bottomrule
\end{tabular}
\end{table}

\begin{table}[t]
\centering
\caption{\textbf{Replica} (\%). Baseline numbers from Gaga~\cite{lyu2024gaga}.}
\label{tab:replica}
\setlength{\tabcolsep}{3.5pt}
\begin{tabular}{@{}llccc@{}}
\toprule
Method & Masks & mIoU & Prec. & Recall \\
\midrule
Gaussian Grouping~\cite{ye2023gaussiangrouping} & EntitySeg & 35.90 & 14.07 & 31.57 \\
Gaga~\cite{lyu2024gaga} & EntitySeg & 41.45 & \textbf{59.74} & 45.59 \\
OP2GS & EntitySeg & \textbf{49.09} & 37.91 & \textbf{55.45} \\
\midrule
Gaussian Grouping~\cite{ye2023gaussiangrouping} & SAM & 21.76 & 25.00 & 19.72 \\
Gaga~\cite{lyu2024gaga} & SAM & 46.21 &\textbf{39.53} & 50.62 \\
OP2GS & SAM & \textbf{54.33} & 37.29 & \textbf{61.24} \\
\bottomrule
\end{tabular}
\end{table}

We first compare OP2GS with the state of the art on four benchmarks (\cref{sec:exp_sota}).
\cref{sec:exp_capability} then measures the capabilities that follow from the representation: frozen-asset segmentation, the ground-truth-mask upper bound, and convergence. \cref{sec:exp_ablation} ablates the design.

\subsection{Datasets, Protocols, and Implementation}
\label{sec:exp_setting}
\paragraph{ScanNet++~\cite{yeshwanth2023scannetpp}.}
We use the 50 validation scenes with the frame selection of OpenSplat3D~\cite{piekenbrinck2025opensplat3d} (at most 300 DSLR frames per scene).
Following the official setting of OpenSplat3D, the Gaussians are initialized from the point cloud provided with each scene.
We evaluate class-agnostic and class-aware 3D instance segmentation on the mesh vertices with the official ScanNet++ evaluation and ignore masks, pooling all 50 scenes into one evaluation (AP, AP$_{50}$, AP$_{25}$); each vertex takes the label of its nearest Gaussian with $\sigma^{*} > \tau$.
For the class-aware setting, every predicted instance is assigned one of the 84 instance classes with MasQCLIP~\cite{xu2023masqclip} features aggregated over its five most visible views.

\paragraph{ScanNet~\cite{dai2017scannet}.}
We follow the protocol of OpenGaussian~\cite{wu2024opengaussian} and InstanceGaussian~\cite{li2025instancegaussian} on ten scenes.
Gaussians are initialized at the mesh vertices and neither densified nor pruned.
We report 3D instance mIoU and mAcc at IoU thresholds 0.25 and 0.5, averaged over scenes.

\paragraph{Replica~\cite{straub2019replica}.}
We follow the protocol of Gaga~\cite{lyu2024gaga} on eight scenes: COLMAP poses, training views every 5th frame, and mIoU / precision / recall of the instance maps rendered on the held-out views.
On Replica and Messy Rooms, 2D instance maps use the sum readout of \cref{sec:primitive} (the max readout is compared in the supplementary).

\paragraph{Messy Rooms~\cite{bhalgat2023contrastivelift}.}
Eight synthetic scenes with 25 to 500 objects each, with the train/test split of the benchmark~\cite{bhalgat2023contrastivelift}: the last 20\% of the frames of every scene are held out, and PQ$^\text{scene}$ is reported on them.
Each predicted instance is classified as thing or stuff by a vote of the provided Detic semantic maps over the training views.

\paragraph{Masks.}
Unless stated otherwise, 2D masks come from SAM ViT-H~\cite{sam_Kirillov_2023_ICCV}; we additionally report OP2GS with entity masks from CropFormer~\cite{qi2023entityseg} (denoted \emph{EntitySeg}), using the same pipeline.
Implementation details and the training schedule are given in the supplementary.

\subsection{Comparison with the State of the Art}
\label{sec:exp_sota}

\begin{figure*}[t]
  \centering
  \includegraphics[width=\linewidth]{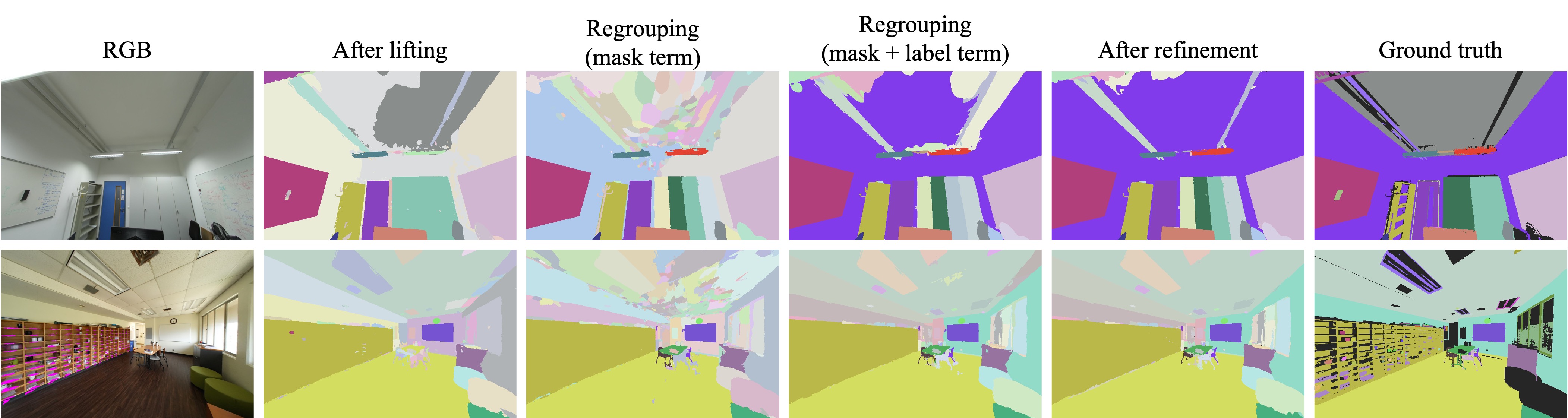}
  \caption{\textbf{Instance maps after each stage} on two ScanNet++ scenes. From left: input view; labels after association and lifting (\cref{sec:assoc}); after regrouping with the mask term only; after regrouping with the mask and label terms (\cref{eq:edge}); after refinement and pruning (\cref{sec:refine}); ground truth. A predicted instance that matches a ground-truth object (IoU $> 0.5$) takes its color, unmatched predictions are shown in pale colors, and black marks pixels without a label. Most ground-truth objects of the second scene are small items on the shelves, which no stage separates.}
  \label{fig:steps}
\end{figure*}

\paragraph{ScanNet++.}
On the 50 validation scenes (\cref{tab:scannetpp}), OP2GS obtains the highest AP with either mask source: +1.0 AP over OpenSplat3D with SAM masks, where OP2GS has slightly lower AP$_{50}$ and AP$_{25}$, and +4.8 AP with EntitySeg masks, where it leads on all metrics. Replacing SAM by EntitySeg masks improves OP2GS by 2.9 AP, whereas OpenSplat3D loses 0.9 AP with the same masks, so OP2GS benefits from better masks, whereas OpenSplat3D does not. Qualitative comparisons are given in the supplementary.

\paragraph{ScanNet.}
On ScanNet (\cref{tab:scannet}), OP2GS exceeds InstanceGaussian in mIoU with both SAM and EntitySeg masks, without any post-processing, whereas InstanceGaussian applies MLP smoothing; with SAM masks it is higher in mAcc@0.25 by a large margin of 7.7 points, while its mAcc@0.5 is 3.0 points lower.

\paragraph{Replica.}
OP2GS improves over Gaga by 8.1 mIoU under the same protocol (\cref{tab:replica}), with 10.6 points higher recall but lower precision.
Gaga with EntitySeg masks has the highest precision, but its mIoU is 12.9 points below OP2GS.

\paragraph{Messy Rooms.}
OP2GS reaches the best mean PQ$^\text{scene}$ (\cref{tab:messy}) and the best score on six of the eight scenes, including the 500-object corridor, while UniC-Lift stays ahead on the 25- and 500-object room.

\subsection{Capabilities of the Representation}
\label{sec:exp_capability}

\begin{table}[t]
\centering
\caption{\textbf{Messy Rooms}, PQ$^\text{scene}$ (\%) on the test frames; column headers give the number of objects per scene. Baseline numbers are taken from UniC-Lift~\cite{dhiman2026uniclift}. Unified-Lift and OP2GS use SAM masks.}
\label{tab:messy}
\setlength{\tabcolsep}{1.6pt}
\footnotesize
\begin{tabular}{@{}lccccccccc@{}}
\toprule
 & \multicolumn{4}{c}{Old room} & \multicolumn{4}{c}{Large corridor} & \\
\cmidrule(lr){2-5}\cmidrule(lr){6-9}
Method & 25 & 50 & 100 & 500 & 25 & 50 & 100 & 500 & Mean \\
\midrule
OmniSeg3D-GS~\cite{ying2024omniseg3d} & 80.1 & 72.4 & 61.4 & 46.8 & 74.9 & 79.6 & 63.9 & 48.5 & 66.0 \\
Unified-Lift~\cite{zhu2025unifiedlift} & 79.1 & 72.2 & 65.9 & 53.9 & 77.0 & 78.9 & 70.7 & 54.1 & 69.0 \\
UniC-Lift~\cite{dhiman2026uniclift} & \textbf{86.0} & 79.1 & 70.8 & \textbf{57.4} & 76.5 & 72.3 & 73.0 & 56.7 & 71.5 \\
\midrule
OP2GS & 78.8 & \textbf{86.3} & \textbf{80.3} & 53.6 & \textbf{77.1} & \textbf{82.0} & \textbf{75.8} & \textbf{57.0} & \textbf{73.9} \\
\bottomrule
\end{tabular}
\end{table}

\begin{table}[t]
\centering
\caption{\textbf{Stage-wise contribution}, ScanNet++ 50 scenes, SAM masks (AP / AP$_{50}$ / AP$_{25}$, \%). ``Unfiltered'' uses all Gaussians at readout; ``$\sigma^{*}{>}0.1$'' keeps confident ones. Before occupancy learning, $\sigma^{*}$ equals $\sigma$.}
\label{tab:stages}
\setlength{\tabcolsep}{3pt}
\small
\begin{tabular}{@{}lcc@{}}
\toprule
Stage & Unfiltered & $\sigma^{*} > 0.1$ \\
\midrule
Association + lifting (\cref{sec:assoc}) & 11.7/26.0/50.3 & 11.9/26.2/50.3 \\
+ occupancy learning (\cref{sec:occupancy}) & 11.7/26.0/50.2 & 13.2/28.5/51.6 \\
+ regrouping, mask term & 17.2/31.6/46.6 & 17.4/31.7/46.9 \\
+ label term (\cref{eq:edge}) & 19.2/36.2/54.3 & 19.4/36.2/54.7 \\
+ refinement (\cref{sec:refine}) & 19.7/36.5/54.5 & \textbf{20.2/37.1/54.7} \\
\bottomrule
\end{tabular}
\end{table}

\begin{figure}[t]
  \centering
  \includegraphics[width=\linewidth]{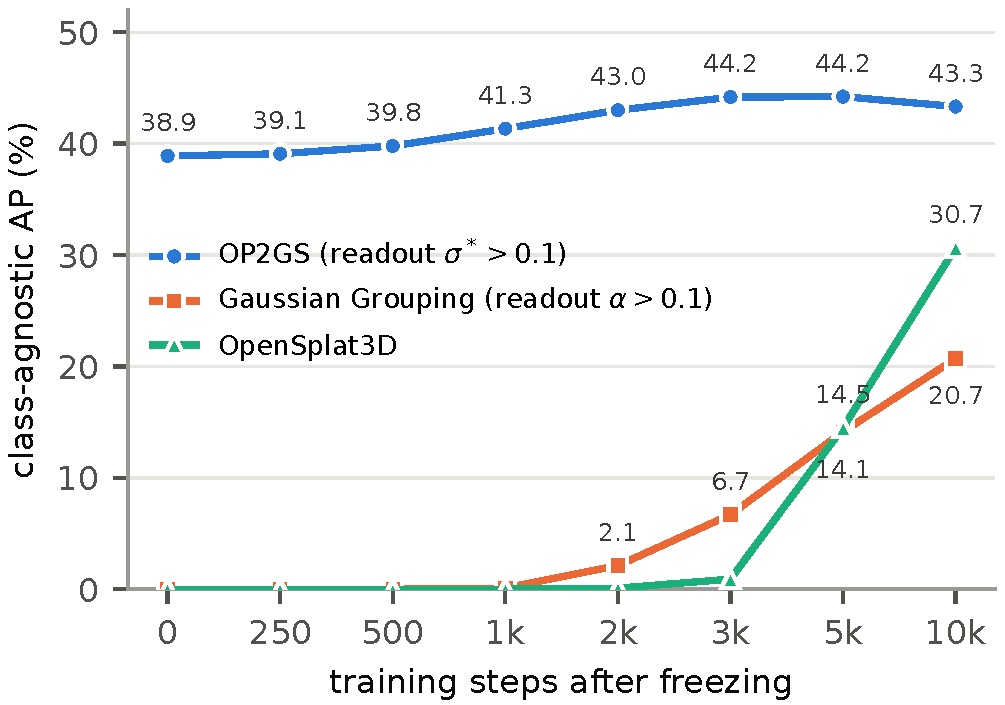}
  \caption{\textbf{Convergence on frozen assets.} Six randomly selected ScanNet++ scenes, one frozen RGB-only 3DGS, ground-truth masks; each method trains only its segmentation parameters. The end points are pooled over these six scenes and are not comparable to \cref{tab:gtupper}, which trains jointly on all 50 scenes.}
  \label{fig:convergence}
\end{figure}

\begin{table}[t]
\centering
\caption{\textbf{Upper bound with ground-truth masks}, ScanNet++ 50 scenes (AP / AP$_{50}$ / AP$_{25}$, \%). All methods receive the same id-consistent ground-truth 2D masks, which isolates the capacity of each representation from mask quality and association. ``Extra'': what each method stores per Gaussian for segmentation. All methods are trained jointly with their full 30k-iteration schedule, unlike the frozen 10k-iteration setting of \cref{fig:convergence}; OP2GS lifts the ground-truth ids and trains $\sigma^{*}$, without regrouping.}
\label{tab:gtupper}
\setlength{\tabcolsep}{2pt}
\small
\begin{tabular}{@{}llc@{}}
\toprule
Method & Extra per Gaussian & AP/AP$_{50}$/AP$_{25}$ \\
\midrule
OpenSplat3D~\cite{piekenbrinck2025opensplat3d} & 8-d feature & 27.9/59.0/85.7 \\
Gaussian Grouping~\cite{ye2023gaussiangrouping} & 16-d enc.\ + classifier & 39.1/\textbf{77.9}/\textbf{93.6} \\
OP2GS & \textbf{1 scalar + 1 integer} & \textbf{42.0}/77.6/93.0 \\
\bottomrule
\end{tabular}
\end{table}

\paragraph{Frozen assets and convergence.}
The pipeline also runs on a completely frozen 3DGS, where only $\sigma^{*}$ and the labels change and pruning is replaced by setting $\sigma^{*}$ to zero.
OP2GS loses 0.3 AP with SAM masks and 0.4 AP with EntitySeg masks in this setting; OpenSplat3D under the same protocol is reported alongside in \cref{tab:frozen}.
On frozen assets with ground-truth masks (\cref{fig:convergence}), voting alone already gives 38.9 AP before any $\sigma^{*}$ gradient step and OP2GS reaches its best accuracy within 3k iterations, whereas the feature-based baselines start from zero and converge slowly, which makes OP2GS an efficient way to label existing 3DGS assets. Since voting gives OP2GS an initialization the baselines cannot use, this measures how quickly each representation absorbs given labels, not optimization speed.

\paragraph{Upper bound with ground-truth masks.}
With the same id-consistent ground-truth masks for all methods, which removes mask quality and association from the comparison, OP2GS reaches a higher AP upper bound than both feature-based methods with two values per Gaussian instead of an 8- or 16-d feature (\cref{tab:gtupper}).

\begin{table}[!h]
\centering
\caption{\textbf{Plug-and-play on frozen 3DGS assets}, ScanNet++ 50 scenes (AP / AP$_{50}$ / AP$_{25}$, \%). \emph{Frozen}: only the segmentation parameters are trained on the same RGB-only 3DGS (30k iterations); all other parameters stay bitwise identical.}
\label{tab:frozen}
\setlength{\tabcolsep}{3pt}
\small
\begin{tabular*}{\linewidth}{@{\extracolsep{\fill}}llcc@{}}
\toprule
Method & Masks & Joint & Frozen \\
\midrule
OpenSplat3D & SAM & 19.2/37.4/56.9 & 17.9/36.0/56.0 \\
OP2GS & SAM & 20.2/37.1/54.7 & \textbf{19.9/36.7/54.5} \\
OpenSplat3D & EntitySeg & 18.3/36.3/55.2 & 18.9/35.8/53.8 \\
OP2GS & EntitySeg & 23.1/40.9/56.8 & \textbf{22.7/40.0/56.2} \\
\bottomrule
\end{tabular*}
\end{table}

\subsection{Ablations}
\label{sec:exp_ablation}


\paragraph{Ablation on pipeline stages.}
\cref{tab:stages} and \cref{fig:steps} separate the contribution of each stage. Occupancy learning keeps the labels, but reading them out with $\sigma^{*}$ instead of $\sigma$ raises AP from 11.9 to 13.2. Regrouping with the mask term of \cref{eq:edge} alone (multi-view mask agreement only) raises the unfiltered AP from 11.7 to 17.2; adding the label term, the prior that keeps regions of the same lifted label together, reaches 19.2; refinement adds +0.5.

\begin{table}[t]
\centering
\caption{\textbf{Dual vs.\ single opacity.} ScanNet++ 50 scenes; GT: id-consistent ground-truth masks, setting of \cref{tab:gtupper}. Readout keeps Gaussians with $\sigma^{*}>0.1$ (\ie, $\sigma>0.1$ for the single-opacity variant). PSNR on held-out views of the SAM-mask models (default, with pruning); first row: RGB-only training of the same scenes.}
\label{tab:dual}
\setlength{\tabcolsep}{2pt}
\small
\begin{tabular}{@{}lcccc@{}}
\toprule
 & SAM masks & GT masks & PSNR & Frozen? \\
\midrule
RGB only & -- & -- & 23.44 & -- \\
Single ($\sigma^{*}{\equiv}\sigma$) & 19.1/35.6/53.5 & 34.0/71.1/92.8 & 22.73 & no \\
Dual (OP2GS) & \textbf{20.2/37.1/54.7} & \textbf{42.0/77.6/93.0} & \textbf{23.40} & yes \\
\bottomrule
\end{tabular}
\end{table}

\paragraph{Dual vs.\ single opacity.}
\cref{tab:dual} compares the dual design with a single opacity that serves both roles. With SAM masks the single-opacity variant reaches 19.1 AP, the level of OpenSplat3D$^\dagger$ (19.2), so the pipeline for feature-free Gaussians alone recovers the baseline; the dual opacity contributes 1.1 AP with SAM masks and 8.0 AP with ground-truth masks, where association and regrouping no longer dominate the error.
Independently of the margin, it is what makes the frozen setting possible: a single opacity cannot be supervised without changing the reconstruction (0.67\,dB PSNR lost), whereas the dual design renders within 0.04\,dB of the RGB-only model and on frozen assets leaves the appearance parameters bitwise identical (\cref{tab:frozen}).
\begin{table}[t]
\centering
\caption{\textbf{Readout threshold}, ScanNet++ 50 scenes, SAM masks (AP / AP$_{50}$ / AP$_{25}$, \%). ``Kept'': the fraction of the Gaussians of the final model (after pruning for the default OP2GS) that carry an instance at readout, \ie, pass the $\sigma^{*}$ threshold for OP2GS, or are assigned to a cluster rather than noise by HDBSCAN for OpenSplat3D. Last block: filtering by the lifting vote share (model without pruning, so that every Gaussian keeps its vote).}
\label{tab:readout}
\setlength{\tabcolsep}{2pt}
\small
\begin{tabular}{@{}llcc@{}}
\toprule
Method & Readout & Kept & AP/AP$_{50}$/AP$_{25}$ \\
\midrule
OpenSplat3D$^\dagger$ & HDBSCAN clusters & \textbf{38\%} & 19.2/37.4/56.9 \\
\midrule
OP2GS & unfiltered & 100\% & 19.7/36.5/54.5 \\
OP2GS & $\sigma^{*}>0.1$ (default) & \textbf{81\%} & 20.2/37.1/54.7 \\
OP2GS & $\sigma^{*}>0.3$ & 67\% & 20.1/37.1/54.7 \\
OP2GS & $\sigma^{*}>0.5$ & 59\% & 20.0/37.1/54.7 \\
\midrule
OP2GS w/o pruning & unfiltered & 100\% & 19.2/36.2/54.3 \\
OP2GS w/o pruning & vote share $>0.3$ & 87\% & 19.4/36.2/54.3 \\
OP2GS w/o pruning & $\sigma^{*}>0.1$ & 63\% & 20.2/37.1/54.7 \\
\bottomrule
\end{tabular}
\end{table}

\paragraph{Readout.}
\cref{tab:readout} varies the readout threshold. The readout discards only Gaussians the model is confident belong to no object ($\sigma^{*}\le0.1$) and keeps 81\% of them, whereas the clustering of OpenSplat3D assigns only 38\% of its Gaussians to an instance and leaves the rest as noise; the threshold is not critical.
Filtering by the lifting vote share instead is almost the same as not filtering, whereas $\sigma^{*}$ adds 0.8 AP (last block).
The representation is also cheap: an average ScanNet++ scene (827k Gaussians) needs 6.6\,MB of segmentation parameters against 26\,MB (8-d) or 53\,MB (16-d) for a feature, and no clustering at readout.
The two regrouping steps, the $\sigma^{*}$ weighting of the label term, the number of supervised labels $M$, the refinement targets, and $\sigma^{*}$ vs.\ the lifting vote share are ablated in the supplementary.
\endgroup

\section{Conclusion}
\label{sec:conclusion}
We have presented OP2GS, a Gaussian primitive that carries its instance membership in its own parameters: an integer identity and an instance opacity $\sigma^{*}$ that is rasterized by the same renderer as the appearance opacity.
Around it we build a pipeline for feature-free Gaussians in which a discrete step corrects the labels and a continuous step trains $\sigma^{*}$.
With one scalar and one integer per Gaussian, OP2GS matches or improves on feature-based methods on four benchmarks, attains a higher ground-truth-mask AP upper bound than the feature-based methods we compared, runs on frozen 3DGS assets with essentially no loss in accuracy, and converges within a few thousand iterations.

\paragraph{Limitations and future work.}
OP2GS carries no feature and relies on 2D masks and their multi-view agreement: pieces of an object that rarely co-occur in a view cannot always be aggregated.
Future work will combine OP2GS with off-the-shelf semantic features from foundation models.

{
    \small
    \bibliographystyle{ieeenat_fullname}
    \bibliography{main}
}

\clearpage
\maketitlesupplementary
\appendix
\setcounter{table}{0}\setcounter{figure}{0}\setcounter{equation}{0}
\renewcommand{\thetable}{S\arabic{table}}
\renewcommand{\thefigure}{S\arabic{figure}}
\renewcommand{\theequation}{S\arabic{equation}}
\renewcommand{\bibliography}[1]{}\renewcommand{\bibliographystyle}[1]{}
This supplementary material gives the protocol details (Sec.~\ref{sec:supp_protocol}), the details and ablations of the association, superpoints and regrouping (Sec.~\ref{sec:supp_assoc}) and of refinement and readout (Sec.~\ref{sec:supp_refine}), qualitative comparisons and an interactive correction (Sec.~\ref{sec:supp_qualitative}), and the convergence protocol (Sec.~\ref{sec:supp_convergence}). The qualitative figures of all four benchmarks and the per-scene results of ScanNet++, ScanNet and Replica follow at the end; per-scene results of Messy Rooms are given in Tab.~\ref{tab:messy} of the main paper.

\section{Protocol Details}
\label{sec:supp_protocol}

\paragraph{ScanNet++.}
Initialization follows OpenSplat3D (750k initial points sampled from the provided point cloud, half resolution), and every predicted instance receives confidence 1 in the evaluation.
OpenSplat3D$^\dagger$ was re-run with its official code on the same frames and masks and evaluated with the same script.
Its published class-agnostic result without post-processing matches our re-run (Tab.~\ref{tab:scannetpp}). Its official export additionally post-processes the predictions on the mesh, by voting the labels within the superpoints of the released mesh segmentation and smoothing them over the mesh vertices; we report all methods without this post-processing.
For the class-aware setting, each predicted 3D instance is cropped at three scales in its five training views with the highest product of visible fraction and rendered area (as for ScanNet below) before the MasQCLIP~\cite{xu2023masqclip} classification; the same procedure is applied to the baselines.
Per-scene results are listed in \cref{tab:supp_perscene_spp}.

\begin{figure*}[t]
  \centering
  \includegraphics[width=\linewidth]{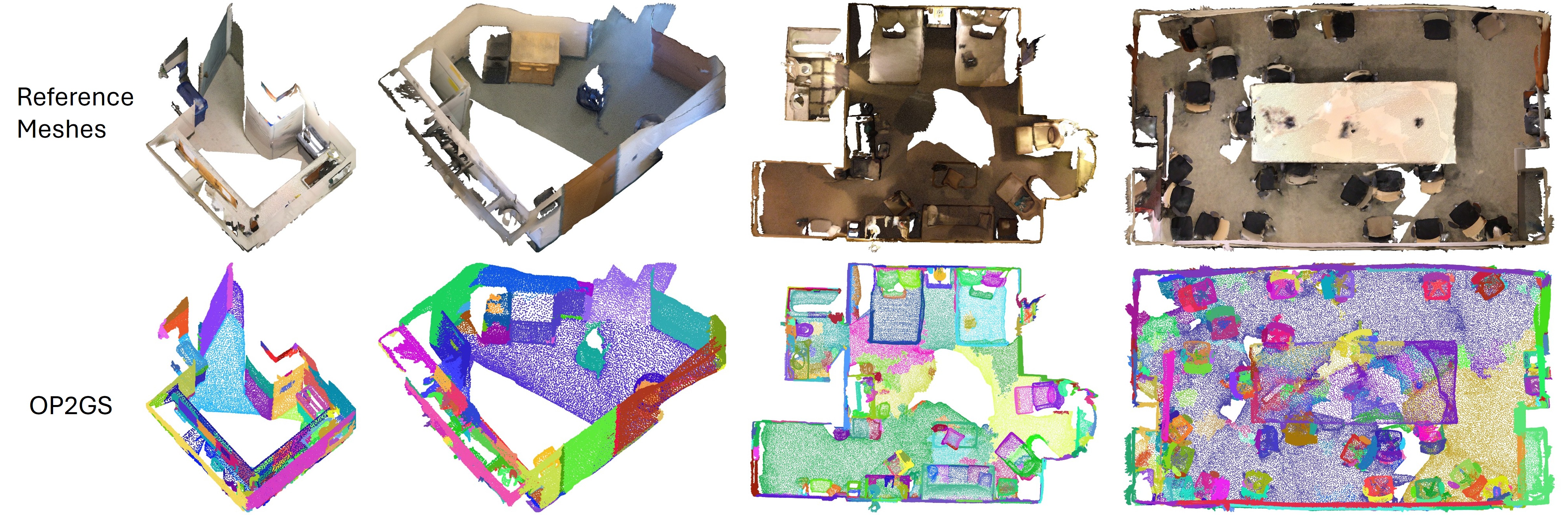}
  \caption{\textbf{Class-agnostic 3D instance segmentation of OP2GS on ScanNet.} Top: reference meshes; bottom: the predicted instances on the mesh vertices ($\sigma^*{>}0.1$ readout), one random color per instance. Ceilings are removed for visibility.}
  \label{fig:supp_scannet_overview}
\end{figure*}

\begin{figure*}[t]
  \centering
  \includegraphics[width=\linewidth]{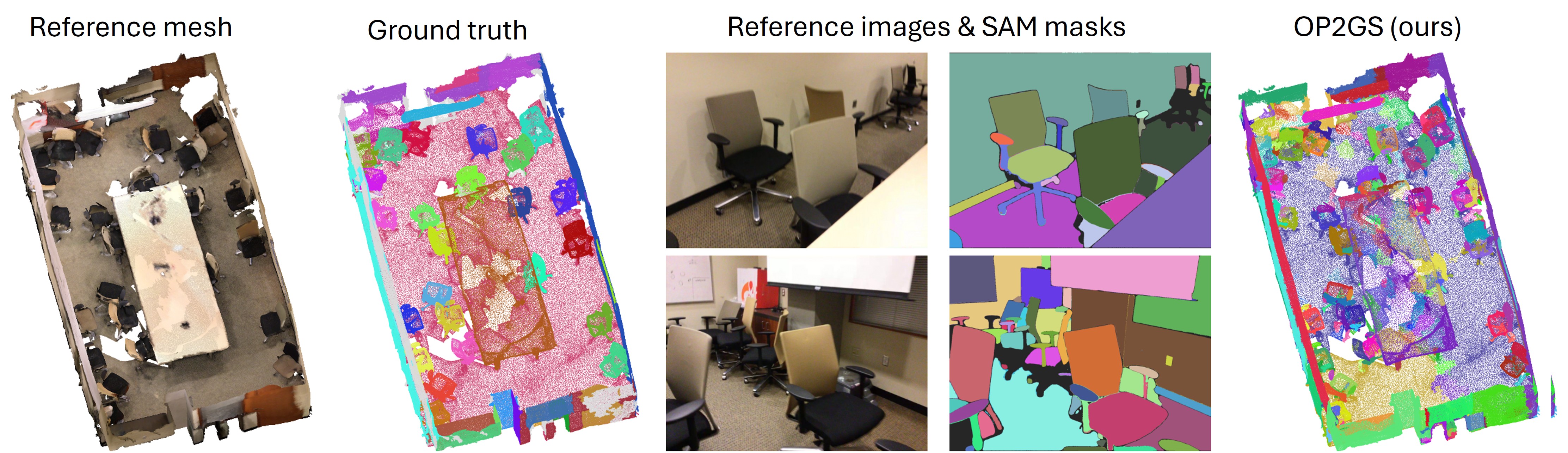}
  \caption{\textbf{Failure case on ScanNet} (scene0140\_00, 43.3 mIoU with SAM masks). Two training views with their SAM masks are shown in the middle. SAM splits each chair into its back, seat, armrests and base, and splits the floor into separate parts, shown in different colors within each view; OP2GS inherits part of this fragmentation. With EntitySeg masks the same scene reaches 61.1 mIoU instead of 43.3 (Tab.~\ref{tab:supp_perscene_scannet}), which confirms that the fragmentation is inherited from the mask source rather than from the representation.}
  \label{fig:supp_failure}
\end{figure*}

\paragraph{ScanNet.}
Following OpenGaussian, the Gaussians are initialized at the vertices of the reference mesh and stay frozen in position, without densification or pruning, so that the $i$-th Gaussian is the $i$-th vertex and every vertex receives a prediction; we use the released evaluation code of OpenGaussian / InstanceGaussian.
Fig.~\ref{fig:supp_scannet_overview} shows the class-agnostic instances of OP2GS on four scenes, and \cref{tab:supp_perscene_scannet} lists the per-scene results.
Fig.~\ref{fig:supp_failure} shows a failure case.

\paragraph{Open-vocabulary semantics on ScanNet.}
OP2GS attaches no language features to Gaussians, but every instance can be rendered as a 2D mask in any view, from which a language feature is computed. \cref{tab:supp_openvocab} reports semantic segmentation on the OpenGaussian protocol with bare class names as queries.
The published CLIP results use the released LangSplat features, \ie, CLIP features of SAM ``whole'' masks. InstanceGaussian trains on the SAM ``whole'' masks and is evaluated on the same masks, so its instances share the granularity of the evaluated masks, which favors it independently of the segmentation quality. Moreover, the views are chosen by the largest mask area alone, which can select views where most of the object is occluded or cut off.

\begin{figure*}[p]
  \centering
  \includegraphics[width=0.9\linewidth]{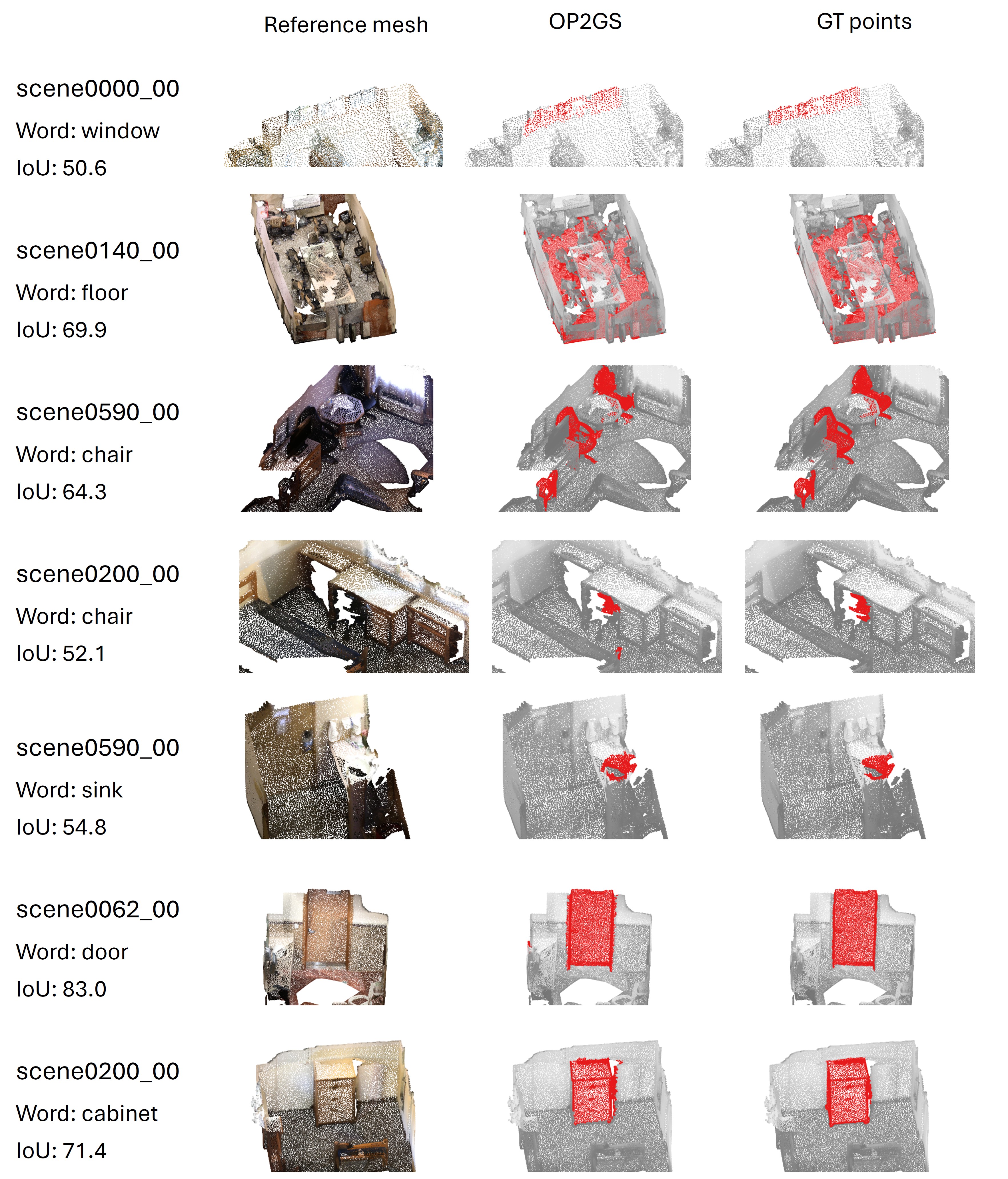}
  \caption{\textbf{Open-vocabulary queries on ScanNet (OP2GS).} Each row shows a close-up of one scene and one class query; red marks the mesh vertices assigned to the queried class (middle) and the ground-truth vertices of that class (right), with the per-scene class IoU above. Instances are classified with MasQCLIP as in Tab.~\ref{tab:supp_openvocab}.}
  \label{fig:supp_ov_scannet}
\end{figure*}

We therefore use MasQCLIP~\cite{xu2023masqclip}, which encodes an image together with an instance mask, and select views as in OpenSplat3D~\cite{piekenbrinck2025opensplat3d} (\cref{eq:ov_view}). For instance $k$ and view $v$,
\begin{equation}
  \begin{gathered}
  w_k(v) = f_k(v)\, A_k(v), \\
  c_k = \arg\max_{c} \Big\langle \sum_{v \in V_k} \mathbf{e}_k(v),\, \mathbf{t}_c \Big\rangle,
  \end{gathered}
  \label{eq:ov_view}
\end{equation}
where $f_k(v)$ is the fraction of the Gaussians of $k$ visible in $v$, $A_k(v)$ its rendered mask area, $V_k$ the five views with the largest $w_k(v)$, $\mathbf{e}_k(v)$ the MasQCLIP feature of the mask (averaged over three crop scales; the sum over $V_k$ equals the average up to a constant factor) and $\mathbf{t}_c$ the text feature of class $c$.
We apply the same procedure, with the same Gaussians, views and language model, to the released instances of InstanceGaussian$^\dagger$, so that the two rows differ only in the instance partition; OP2GS is ahead by 6.0, 5.6 and 7.6 mIoU on the three class sets.
Fig.~\ref{fig:supp_ov_scannet} visualizes the queried classes on the reference mesh.

\begin{table}[t]
\centering
\caption{\textbf{Open-vocabulary semantic segmentation on ScanNet}, OpenGaussian protocol, 10 scenes, 19 / 15 / 10 classes (mIoU / mAcc, \%). Rows with CLIP are taken from InstanceGaussian~\cite{li2025instancegaussian} and use the released LangSplat features (OpenCLIP ViT-B/16 of SAM ``whole'' masks). MasQCLIP rows use the rendered instance masks and the identical procedure for both methods; $^\dagger$our evaluation of the released InstanceGaussian instances.}
\label{tab:supp_openvocab}
\setlength{\tabcolsep}{2.5pt}
\footnotesize
\begin{tabular*}{\linewidth}{@{\extracolsep{\fill}}lccc@{}}
\toprule
Method (features) & 19 classes & 15 classes & 10 classes \\
\midrule
OpenGaussian~\cite{wu2024opengaussian} (CLIP) & 24.73/41.54 & 30.13/48.25 & 38.29/55.19 \\
MaskClustering~\cite{yan2024maskclustering} & 34.59/54.32 & 34.00/45.07 & 38.10/56.74 \\
InstanceGaussian~\cite{li2025instancegaussian} (CLIP) & 40.66/54.01 & 42.51/59.15 & 47.94/64.01 \\
\midrule
InstanceGaussian$^\dagger$ (MasQCLIP) & 45.23/66.65 & 46.23/67.65 & 56.93/76.89 \\
OP2GS (MasQCLIP) & \textbf{51.24}/\textbf{70.66} & \textbf{51.87}/\textbf{70.57} & \textbf{64.50}/\textbf{82.07} \\
\bottomrule
\end{tabular*}
\end{table}

\begin{figure*}[p]
  \centering
  \includegraphics[width=0.77\linewidth]{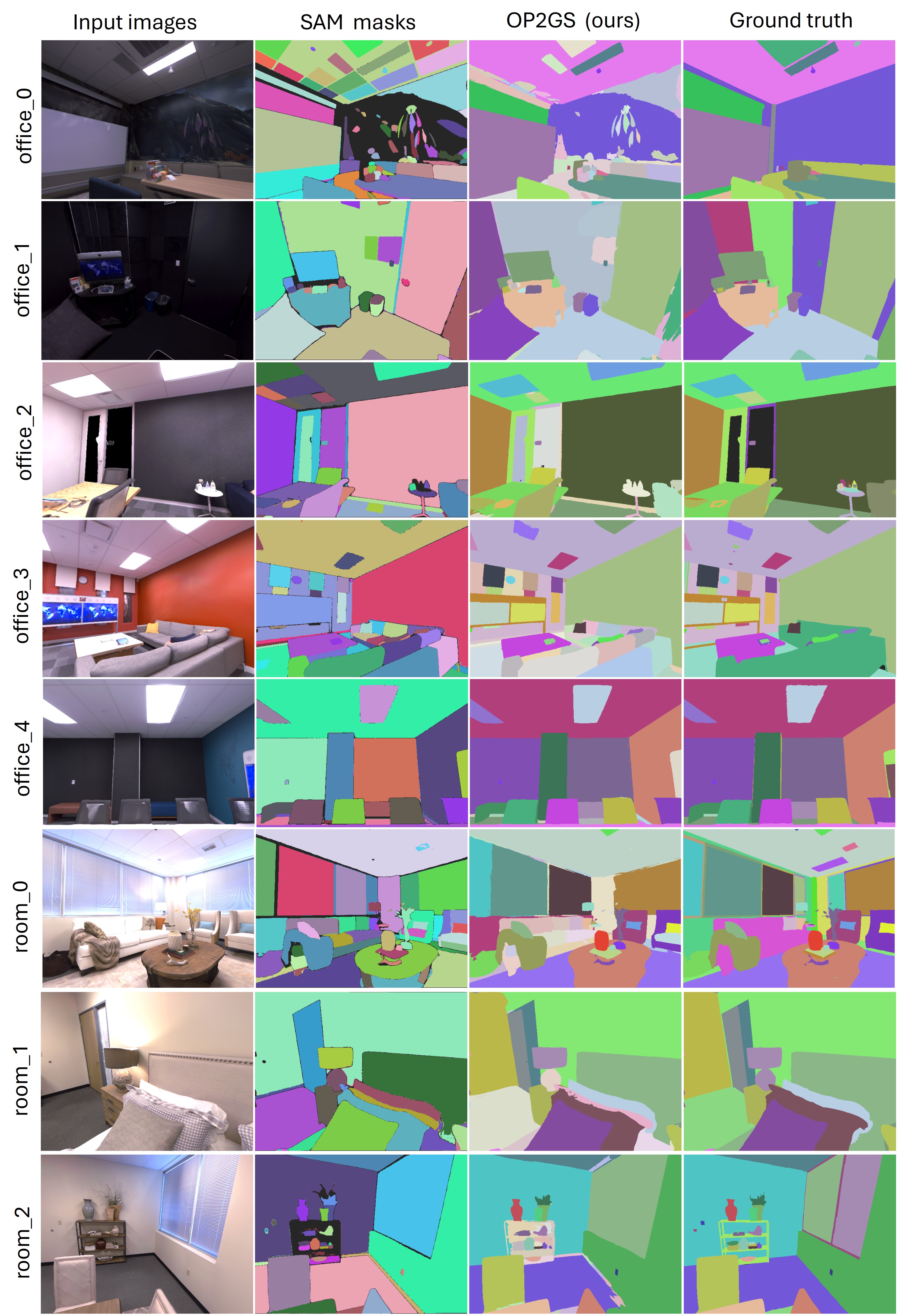}
  \caption{\textbf{Instance segmentation of OP2GS on Replica} (Gaga protocol, one test view per scene). The segmentation largely follows the granularity of the SAM masks: regions that SAM separates in some views are merged by the co-visibility consensus only when enough other views do not separate them, and otherwise remain separate, as for the wall segments in office\_1.}
  \label{fig:supp_replica}
\end{figure*}

\paragraph{Replica.}
We follow the protocol of Gaga~\cite{lyu2024gaga}: training frames 0, 5, \dots, 895 and test frames in between, $640{\times}480$ images, and COLMAP poses for all methods, estimated with the official 3DGS script as in Gaga. Initial points are triangulated with these poses from the training images only, so that the test images do not contribute to the initialization.
Fig.~\ref{fig:supp_replica} shows OP2GS on one test view of each scene, and \cref{tab:supp_perscene_replica} lists the per-scene results.

\begin{figure*}[p]
  \centering
  \includegraphics[width=0.74\linewidth]{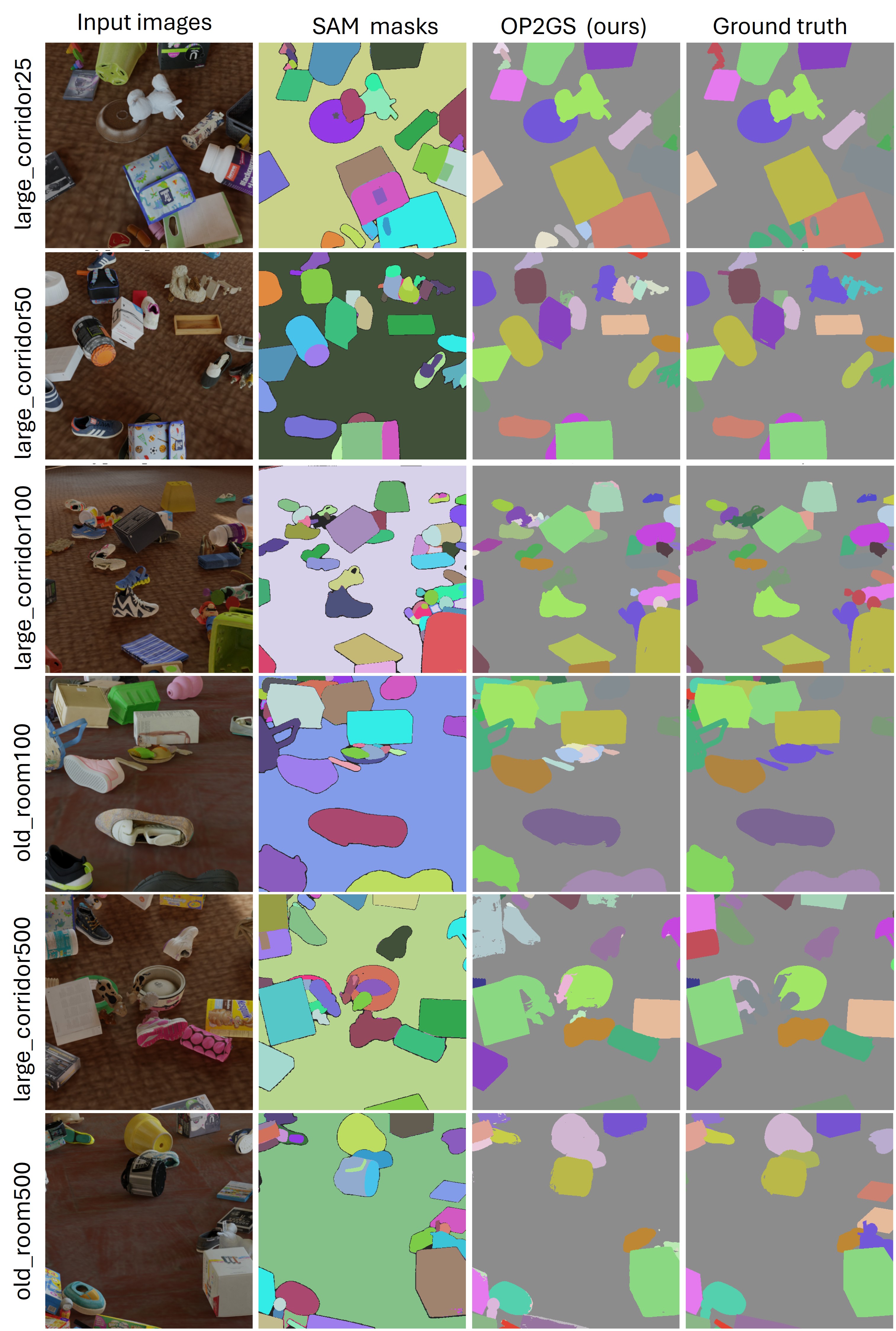}
  \caption{\textbf{Instance segmentation of OP2GS on Messy Rooms} (one test view per scene). Columns: input image, SAM masks of the view (one random color per mask), the rendered instances of OP2GS (sum readout), and the ground truth; gray marks the background.}
  \label{fig:supp_messy}
\end{figure*}

\paragraph{Messy Rooms.}
The baseline numbers of the main table are taken from UniC-Lift~\cite{dhiman2026uniclift}; Unified-Lift classifies the largest segment as background, whereas OP2GS votes with the official Detic semantic maps provided with the benchmark (\cref{sec:exp_setting}).
Camera poses are the provided ground truth; since the benchmark provides no point cloud, initial points are triangulated with these poses from the training frames.
Fig.~\ref{fig:supp_messy} shows one test view of six scenes.

\paragraph{Implementation details.}
2D masks are generated with the default automatic generator of SAM ViT-H; the part level is used only inside the association.
All stages share one schedule and one set of hyperparameters on all datasets (\cref{tab:supp_hparams}).

\paragraph{Single-opacity variant.}
The single-opacity variant of Tab.~\ref{tab:dual} sets $\sigma^{*}{\equiv}\sigma$: the occupancy of Eq.~(\ref{eq:occupancy}) is rendered with the appearance opacity, so $\mathcal{L}_\text{occ}$ updates $\sigma$ in addition to the geometry, and the label term of regrouping, pruning and the readout ($\sigma{>}0.1$) all use $\sigma$. Everything else, including the schedule and the masks, is identical to the full model.

\section{Association, Superpoints and Regrouping}
\label{sec:supp_assoc}
\label{sec:supp_superpoint}

\paragraph{Association.}
Algorithm~\ref{alg:assoc} summarizes the procedure. The mask score $s_m(i)$ of Gaussian $i$ for a mask $m$ in view $t$ is the share of its rendering weight $\alpha_i T_i$ in $t$ that falls inside $m$; the scores of all Gaussians are obtained at once by one rendering pass with constant colors. A group is visible in a view if at least half of its core receives rendering weight there, and it is attributed to the mask of that view that receives most of this weight (provided that mask receives at least half of this weight; otherwise the attribution is undecided). The consensus ignores views that contain a member mask of either group. The containment margin $c$ makes the containment rule of Sec.~\ref{sec:assoc} precise.

\begin{algorithm}[t]
\caption{Training-free mask association}
\label{alg:assoc}
\small
\begin{algorithmic}[1]
\Require trained 3DGS (iteration 20k); SAM masks $\mathcal{M}_t$ of every training view $t$; core threshold $\tau_c{=}0.8$; containment margin $c{=}0.05$; consensus thresholds $\theta = 0.9, 0.85, \dots, 0.7$; minimum views $n_{\min}{=}3$; part-level fraction $\rho{=}0.9$
\Ensure view-consistent masks (one global id per mask)
\Procedure{Associate}{}
  \For{each view $t$ and mask $m \in \mathcal{M}_t$}
    \State $C_m \gets \{ i : s_m(i) \ge \tau_c \}$ \Comment{core}
  \EndFor
  \State $\mathcal{K} \gets \{\{m\}\}$ for all masks \Comment{one group per mask}
  \For{$\theta = 0.9, 0.85, \dots, 0.7$}
    \Repeat
      \State $P \gets \{(A,B) : \Call{Consensus}{A,B} \ge \theta\}$
      \State merge pairs of $P$ in decreasing consensus, each group at most once
    \Until{$P = \emptyset$}
  \EndFor
  \State merge groups whose cores lie in the same part-level mask in $\ge \rho$ of their co-visible views
  \State discard groups observed in fewer than $n_{\min}$ views
  \State \Return a global id per group; rewrite the pixels of every mask with its id
\EndProcedure
\Procedure{Consensus}{$A, B$}
  \State $V \gets$ views where $A$ and $B$ are both visible and attributed, excluding views that contain a member mask of $A$ or $B$
  \State remove from $V$ the views where $A$ or $B$ covers less than $c$ of its attributed mask \Comment{containment}
  \If{$|V| < n_{\min}$} \State \Return $0$ \EndIf
  \State \Return $|\{u \in V : \mathrm{att}(A,u) = \mathrm{att}(B,u)\}| \,/\, |V|$
\EndProcedure
\end{algorithmic}
\end{algorithm}

\paragraph{Replacing the association.}
To isolate the association, we replace its view-consistent masks by the masks of DEVA~\cite{cheng2023deva} in its official automatic mode (SAM with DEVA tracking, as used in Gaussian Grouping~\cite{ye2023gaussiangrouping} and ObjectGS~\cite{zhu2025objectgs}), on the same training frames; all later stages are unchanged and still use the raw SAM masks (\cref{tab:supp_deva}).
DEVA reassigns ids when an object leaves and re-enters the view, so the lifted labels are fragmented (3.8 vs.\ 13.2 AP after lifting). Regrouping recovers most of this gap (16.3 vs.\ 19.4 AP), so the pipeline remains usable with a weaker association; the remaining 2.1 AP of the final model (18.1 vs.\ 20.2 AP) comes from our association.

\begin{table}[t]
\centering
\caption{\textbf{Association source}, ScanNet++ 50 scenes, SAM masks, $\sigma^{*}{>}0.1$ readout (AP / AP$_{50}$ / AP$_{25}$, \%). DEVA is the official release in automatic mode. Only the view-consistent masks used for lifting change; all later stages are identical. Lifting: after occupancy learning; Regrouping: right after regrouping; Final: after refinement.}
\label{tab:supp_deva}
\setlength{\tabcolsep}{3pt}
\small
\begin{tabular*}{\linewidth}{@{\extracolsep{\fill}}lccc@{}}
\toprule
Association & Lifting & Regrouping & Final \\
\midrule
DEVA & 3.8/9.6/28.8 & 16.3/32.1/50.0 & 18.1/34.0/51.1 \\
Ours & \textbf{13.2/28.5/51.6} & \textbf{19.4/36.2/54.7} & \textbf{20.2/37.1/54.7} \\
\bottomrule
\end{tabular*}
\end{table}

\paragraph{One-to-one vs.\ many-to-one targets.}
During occupancy learning, the rendered labels of a view are matched to its SAM masks by Hungarian matching on mask IoU, so each label receives at most one mask.
A label that covers two objects is matched to the mask of one of them, the mask of the other object becomes a negative, and the $\sigma^{*}$ of the whole label drops (\cref{tab:supp_sigma}).
Since the label term of Eq.~(\ref{eq:edge}) is weighted by $\sigma^{*}$, it contributes almost nothing to the edges of such a label, and regrouping can split it by the mask term.
After regrouping, a label should cover one complete object, and one-to-one matching would penalize objects that SAM splits into parts.
The refinement stage therefore uses many-to-one targets, which raise $\sigma^{*}$ again (\cref{tab:supp_sigma}); the pruning and the readout threshold both rely on a high $\sigma^{*}$ for the Gaussians of objects.

\begin{table}[t]
\centering
\caption{\textbf{Mean $\sigma^{*}$ of the Gaussians of a label}, ScanNet++, SAM masks. Mixed: labels that cover two objects after lifting; clean: all other labels.}
\label{tab:supp_sigma}
\setlength{\tabcolsep}{3pt}
\small
\begin{tabular*}{\linewidth}{@{\extracolsep{\fill}}lcc@{}}
\toprule
Labels & \begin{tabular}[b]{@{}c@{}}After occupancy learning\\(one-to-one)\end{tabular} & \begin{tabular}[b]{@{}c@{}}After refinement\\(many-to-one)\end{tabular} \\
\midrule
Mixed & 0.04 & 0.45 \\
Clean & 0.40 & 0.88 \\
\bottomrule
\end{tabular*}
\end{table}

\paragraph{Superpoints and regrouping.}
Superpoints are Felzenszwalb segments~\cite{felzenszwalb2004efficient} on the $k$-NN graph of the centers of visible Gaussians ($\sigma_i \ge 1/255$) and never cross a current label.
A superpoint is visible in a view if at least three of its Gaussians project to pixels whose alpha-composited depth $\sum_i \alpha_i T_i z_i$ ($z_i$: depth of Gaussian $i$) matches the depth of the Gaussian itself within $2\%$.
The affinity of two units (superpoints before region growing, regions after) in a view is the cosine similarity of their SAM-id coverage distributions, weighted by the seen fraction of each unit, and averaged over co-visible views.

\begin{table}[t]
\centering
\caption{\textbf{Instance opacity in the label term} of Eq.~(\ref{eq:edge}) and the gain of refinement (\%). The first two columns are evaluated right after regrouping (before refinement); ``Fixed'' sets $\sqrt{\bar\sigma^*_p\bar\sigma^*_q}{=}1$, \ie, the label prior is $\pm\lambda_l$. ``+ refinement'' is the final model of the main paper. Metrics: AP on ScanNet++, mIoU on ScanNet and Replica, PQ$^\text{scene}$ on Messy Rooms; readout $\sigma^{*}{>}0.1$ on ScanNet++ and ScanNet, sum on Replica and Messy Rooms. $\Delta$: final model minus the fixed label term, \ie, the combined gain of the $\sigma^{*}$ weighting and refinement.}
\label{tab:labelconf}
\setlength{\tabcolsep}{3pt}
\small
\begin{tabular*}{\linewidth}{@{\extracolsep{\fill}}lcccc@{}}
\toprule
Dataset & Fixed $\pm\lambda_l$ & $\times\sqrt{\bar\sigma^*_p\bar\sigma^*_q}$ & + refinement & $\Delta$ \\
\midrule
ScanNet++ & 19.4 & 19.4 & \textbf{20.2} & $+$0.8 \\
ScanNet & 51.31 & 51.47 & \textbf{51.90} & $+$0.59 \\
Replica & 51.50 & 53.08 & \textbf{54.33} & $+$2.83 \\
Messy Rooms & 69.3 & 71.5 & \textbf{73.9} & $+$4.6 \\
\bottomrule
\end{tabular*}
\end{table}

\paragraph{Instance opacity in the label term.}
\cref{tab:labelconf} ablates the instance opacity $\sigma^{*}$ in the label term of Eq.~(\ref{eq:edge}).

\begin{table}[t]
\centering
\caption{\textbf{Training schedule and hyperparameters}, shared by all datasets, except that on ScanNet the Gaussians stay fixed at the mesh vertices, without densification or pruning (Sec.~\ref{sec:supp_protocol}). Values already stated in the main paper are repeated here for completeness. $N$ is the number of Gaussians.}
\label{tab:supp_hparams}
\setlength{\tabcolsep}{6pt}
\small
\begin{tabular*}{\linewidth}{@{\extracolsep{\fill}}ll@{}}
\toprule
Parameter & Value \\
\midrule
\multicolumn{2}{@{}l}{\emph{Training schedule (30k iterations in total)}} \\
RGB-only 3DGS training & 0--20k \\
\quad densification & 0.5k--15k \\
\quad opacity reset & every 3k until 20k \\
Association and lifting & at 20k \\
Occupancy learning & 20k--25k \\
\quad $\sigma^{*}$ init.\ / learning rate & $\sigma^{*}{=}\sigma$ / 0.05 \\
\quad $\lambda_\text{occ}$, labels per step $M$ & 0.1, 3 \\
Regrouping ($\lambda_l$) & at 25k (0.5) \\
Refinement & 25k--30k \\
\quad pruning & at 26k, 27k, 28k \\
Readout threshold $\tau$ & 0.1 \\
Association ($\tau_c, c, \theta, n_\text{min}, \rho$) & Alg.~\ref{alg:assoc} \\
\midrule
\multicolumn{2}{@{}l}{\emph{Superpoints}} \\
Graph & 8-NN, visible centers \\
Edge cost & $1-|n_i^{\top} n_j| + \|c_i - c_j\|$ \\
Felzenszwalb threshold constant & $0.01$ \\
Minimum superpoint size & $N/40000$ \\
\midrule
\multicolumn{2}{@{}l}{\emph{Region growing}} \\
Affinity thresholds (3 stages) & $0.9,\ 0.75,\ 0.6$ \\
\bottomrule
\end{tabular*}
\end{table}

\section{Refinement and Readout}
\label{sec:supp_refine}
\paragraph{Gradient paths of the occupancy loss.}
\cref{tab:supp_grad} changes only where the gradient of $\mathcal{L}_\text{occ}$ flows. In the full pipeline, the effect of the gradient path is mixed with errors of the association and regrouping, so we use the setting of the ground-truth upper bound (Tab.~\ref{tab:gtupper}): labels are lifted from id-consistent ground-truth masks and stay fixed, and the occupancy is trained directly with these masks, without one-to-one matching or regrouping.
Letting $\mathcal{L}_\text{occ}$ also update the geometry through $G_i$ adds 2.3 AP over training $\sigma^{*}$ alone.
Stopping the gradient at $T_i$ (Eq.~(\ref{eq:grad})) changes neither accuracy nor rendering measurably; its purpose is that $\mathcal{L}_\text{occ}$ never reaches the appearance opacity, which makes the frozen setting possible (Tab.~\ref{tab:frozen}).

\begin{table}[t]
\centering
\caption{\textbf{Gradient paths of the occupancy loss}, ScanNet++ 50 scenes, id-consistent ground-truth masks (setting of Tab.~\ref{tab:gtupper}, but with the three pruning steps of the full pipeline), $\sigma^{*}{>}0.1$ readout (AP / AP$_{50}$ / AP$_{25}$, \%); PSNR / SSIM on held-out views. This ablation applies the pruning steps of the full pipeline, which differ from the pruning schedule of Tab.~\ref{tab:gtupper}, so the default row differs slightly from Tab.~\ref{tab:gtupper} (41.5 vs.\ 42.0 AP); the comparison between the three rows is unaffected. PSNR differs by at most 0.02\,dB between the three rows.}
\label{tab:supp_grad}
\setlength{\tabcolsep}{2pt}
\small
\begin{tabular*}{\linewidth}{@{\extracolsep{\fill}}cccc@{}}
\toprule
\multicolumn{2}{c}{Gradient to} & & \\
\cmidrule(r){1-2}
$G_i$ & $T_i$ & AP / AP$_{50}$ / AP$_{25}$ & PSNR / SSIM \\
\midrule
 &  & 39.2 / 75.3 / 92.8 & 23.14 / 0.852 \\
\checkmark &  & \textbf{41.5} / \textbf{77.2} / 93.2 & 23.13 / 0.852 \\
\checkmark & \checkmark & \textbf{41.5} / 77.0 / \textbf{93.5} & 23.12 / 0.851 \\
\bottomrule
\end{tabular*}
\end{table}

\paragraph{Many-to-one targets.}
The refinement stage replaces one-to-one matching by many-to-one targets (\cref{sec:supp_assoc}).
At the start of the refinement stage, every raw SAM mask of every view is assigned to the label that covers the majority of its pixels in the rendered instance map; masks without such a label become negatives for all labels.
The targets stay fixed for the whole stage, and several masks can supervise the same label, so the parts of an object no longer act as negatives for each other.

\paragraph{Pruning.}
The $\sigma^{*}$ threshold for pruning is computed once, at the first pruning step, over the visible Gaussians, and kept fixed afterwards.
The pruned Gaussians are mostly floaters and boundary Gaussians that no object claims.
\cref{tab:supp_prune}: pruning removes 28\% of the Gaussians at a cost of 0.05\,dB PSNR and leaves the AP unchanged.

\begin{table}[t]
\centering
\caption{\textbf{Pruning and rendering quality}, ScanNet++ 50 scenes, SAM masks. Gaussians: mean number per scene after refinement; PSNR / SSIM on held-out views; AP / AP$_{50}$ / AP$_{25}$ (\%) with the $\sigma^{*}{>}0.1$ readout.}
\label{tab:supp_prune}
\setlength{\tabcolsep}{3pt}
\small
\begin{tabular*}{\linewidth}{@{\extracolsep{\fill}}lccc@{}}
\toprule
 & Gaussians & PSNR / SSIM & AP \\
\midrule
RGB-only 3DGS & 827k & 23.44 / 0.857 & -- \\
OP2GS w/o pruning & 827k & 23.45 / 0.857 & 20.2/37.1/54.7 \\
OP2GS & 597k & 23.40 / 0.856 & 20.2/37.1/54.7 \\
\bottomrule
\end{tabular*}
\end{table}

\paragraph{Readout.}
A 2D instance map can be read out in two ways.
The max readout assigns pixel $v$ the label of the single Gaussian with the largest $\alpha^{*}_i T_i$ along the ray and is fast.
The sum readout instead accumulates, for every label $L$, the contributions of all Gaussians of $L$ along the ray, $O_L(v) = \sum_{i:\, l_i = L} \alpha^{*}_i T_i$ (Eq.~(\ref{eq:occupancy})), and assigns $v$ the label with the largest $O_L(v)$; a pixel where every $O_L(v)$ is below $10^{-3}$ is background.
The sum readout gives better results (\cref{tab:supp_readout2d}); the max readout can be used when a good result is needed quickly.
Filtering by the lifting vote share instead of $\sigma^{*}$ is compared in Tab.~\ref{tab:readout} of the main paper.

\begin{table}[t]
\centering
\caption{\textbf{Sum vs.\ max readout} of the final model on the 2D benchmarks (\%). Replica: mIoU / precision / recall; Messy Rooms: PQ$^\text{scene}$ (things / stuff). The sum readout is used in the main paper.}
\label{tab:supp_readout2d}
\setlength{\tabcolsep}{3pt}
\small
\begin{tabular*}{\linewidth}{@{\extracolsep{\fill}}lcc@{}}
\toprule
Readout & Replica & Messy Rooms \\
\midrule
Max & 52.93 / 36.10 / 60.11 & 70.0 (47.4 / 92.5) \\
Sum & \textbf{54.33} / \textbf{37.29} / \textbf{61.24} & \textbf{73.9} (\textbf{53.4} / \textbf{94.4}) \\
\bottomrule
\end{tabular*}
\end{table}

\paragraph{Run-to-run variance.}
We rerun the full pipeline on the 50 ScanNet++ validation scenes with two further random seeds; each run trains the 3DGS from scratch, so the seed changes the sampled initial points and the densification and pruning, \ie, the distribution of the Gaussians, as well as all later stages, including the order in which labels are supervised (\cref{tab:supp_seeds}).
AP is $20.1 \pm 0.2$ over the three runs, and the lowest run (19.8) remains above OpenSplat3D$^\dagger$ (19.2).

\begin{table}[t]
\centering
\caption{\textbf{Run-to-run variance}, ScanNet++ 50 scenes, SAM masks (AP / AP$_{50}$ / AP$_{25}$, \%). Each seed trains the 3DGS from scratch and reruns every stage of the pipeline. Seed 0 is the run reported in the main paper; mean $\pm$ sample standard deviation over the three runs.}
\label{tab:supp_seeds}
\setlength{\tabcolsep}{2pt}
\small
\begin{tabular*}{\linewidth}{@{\extracolsep{\fill}}lcc@{}}
\toprule
Seed & Unfiltered & $\sigma^{*}{>}0.1$ \\
\midrule
0 & 19.7/36.5/54.5 & 20.2/37.1/54.7 \\
1 & 19.4/36.0/54.1 & 19.8/36.8/53.8 \\
2 & 19.8/36.8/53.5 & 20.2/37.1/53.8 \\
\midrule
Mean & 19.6/36.4/54.0 & 20.1/37.0/54.1 \\
Std & 0.2/0.4/0.5 & 0.2/0.2/0.5 \\
\bottomrule
\end{tabular*}
\end{table}

\paragraph{Additional ablations.}
\cref{tab:supp_regroup} adds the components of regrouping one at a time. Region growing alone, which merges superpoints by mask affinity, reaches 17.6 AP; GAEC on superpoints without region growing reaches 19.2 AP; combining both with the mask term only gives 18.0 AP, below GAEC with both terms (19.2 AP), so the label term is the main source of the gain; adding the label term on top gives 20.2 AP.
\cref{tab:supp_M} varies the number of labels supervised per step, $M$; the default $M{=}3$ is a trade-off between training memory and accuracy.

\begin{table}[t]
\centering
\caption{\textbf{Regrouping design}, ScanNet++ 50 scenes, SAM masks, final model after refinement, $\sigma^{*}{>}0.1$ readout (AP / AP$_{50}$ / AP$_{25}$, \%). Region growing merges superpoints by mask affinity alone; the mask and label terms are the two edge weights of GAEC (Eq.~(\ref{eq:edge})); ``const.'': label term $\pm\lambda_l$ without the $\sigma^{*}$ weight.}
\label{tab:supp_regroup}
\setlength{\tabcolsep}{3pt}
\small
\begin{tabular*}{\linewidth}{@{\extracolsep{\fill}}ccccc@{}}
\toprule
Region & \multicolumn{3}{c}{GAEC} & \\
\cmidrule(lr){2-4}
growing & used & mask term & label term & AP / AP$_{50}$ / AP$_{25}$ \\
\midrule
\checkmark &  &  &  & 17.6 / 32.2 / 47.6 \\
 & \checkmark & \checkmark & $\sigma^{*}$ & 19.2 / 35.5 / 52.2 \\
\checkmark & \checkmark & \checkmark &  & 18.0 / 32.5 / 47.3 \\
\checkmark & \checkmark & \checkmark & const. & 20.2 / 37.2 / 54.1 \\
\checkmark & \checkmark & \checkmark & $\sigma^{*}$ & \textbf{20.2 / 37.1 / 54.7} \\
\bottomrule
\end{tabular*}
\end{table}

\begin{table}[t]
\centering
\caption{\textbf{Number of labels supervised per step} $M$ in occupancy learning, ScanNet++ 50 scenes, SAM masks, $\sigma^{*}{>}0.1$ readout (\%). $M{=}3$ is the full model of Tab.~\ref{tab:readout}.}
\label{tab:supp_M}
\setlength{\tabcolsep}{4pt}
\small
\begin{tabular*}{\linewidth}{@{\extracolsep{\fill}}lccc@{}}
\toprule
$M$ & AP & AP$_{50}$ & AP$_{25}$ \\
\midrule
1 & 19.8 & 36.6 & 54.2 \\
2 & 20.1 & 36.9 & 54.7 \\
3 (default) & \textbf{20.2} & \textbf{37.1} & \textbf{54.7} \\
\bottomrule
\end{tabular*}
\end{table}

\section{Qualitative Comparison and Interactive Correction}

\label{sec:supp_qualitative}

\begin{figure*}[t]
  \centering
  \includegraphics[width=0.8\linewidth]{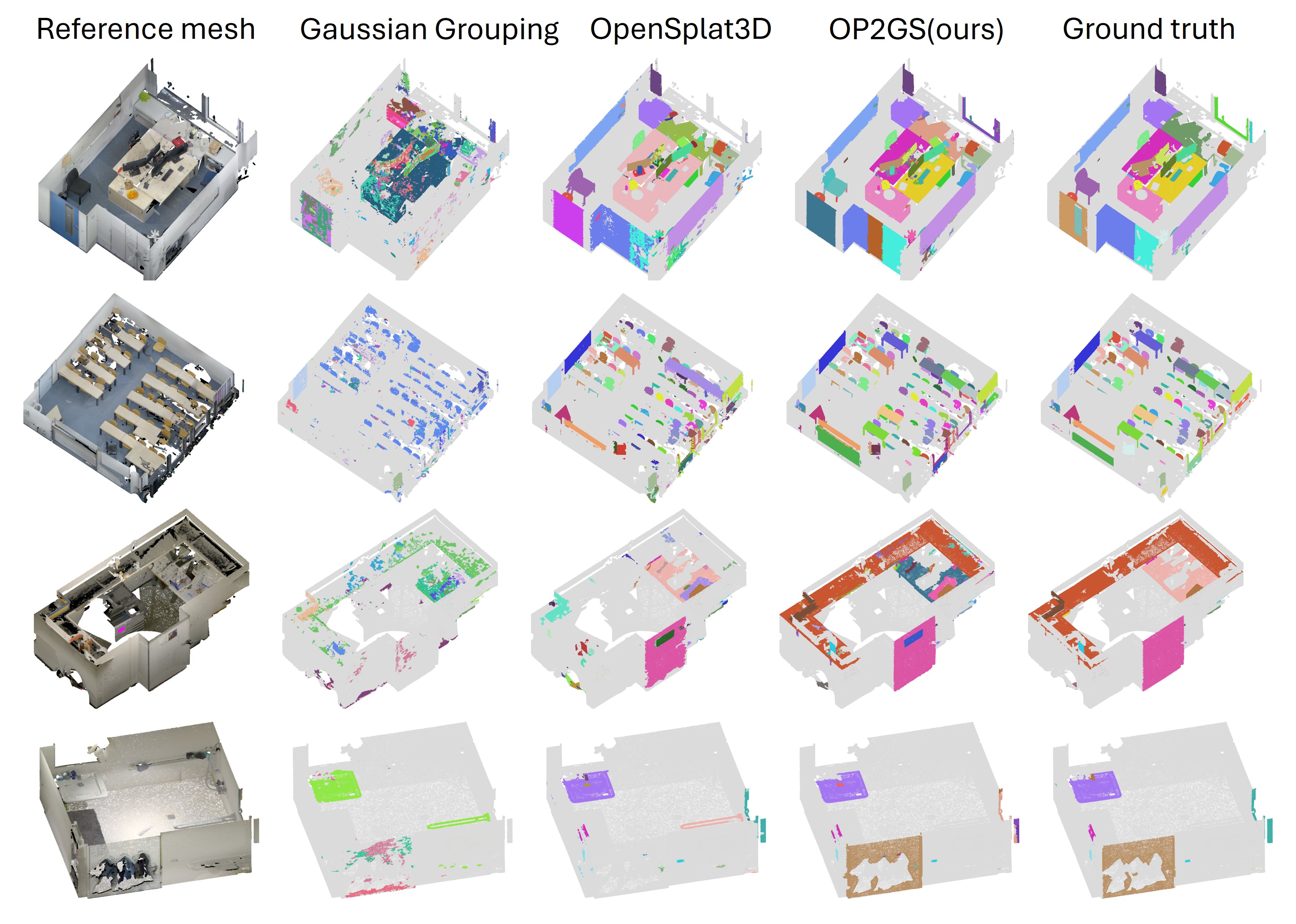}
  \caption{\textbf{Qualitative comparison on ScanNet++.} Predicted instances are transferred to the vertices of the reference mesh; each color denotes one instance.}
  \label{fig:vis_scannetpp}
  \vspace{3mm}
  \includegraphics[width=0.8\linewidth]{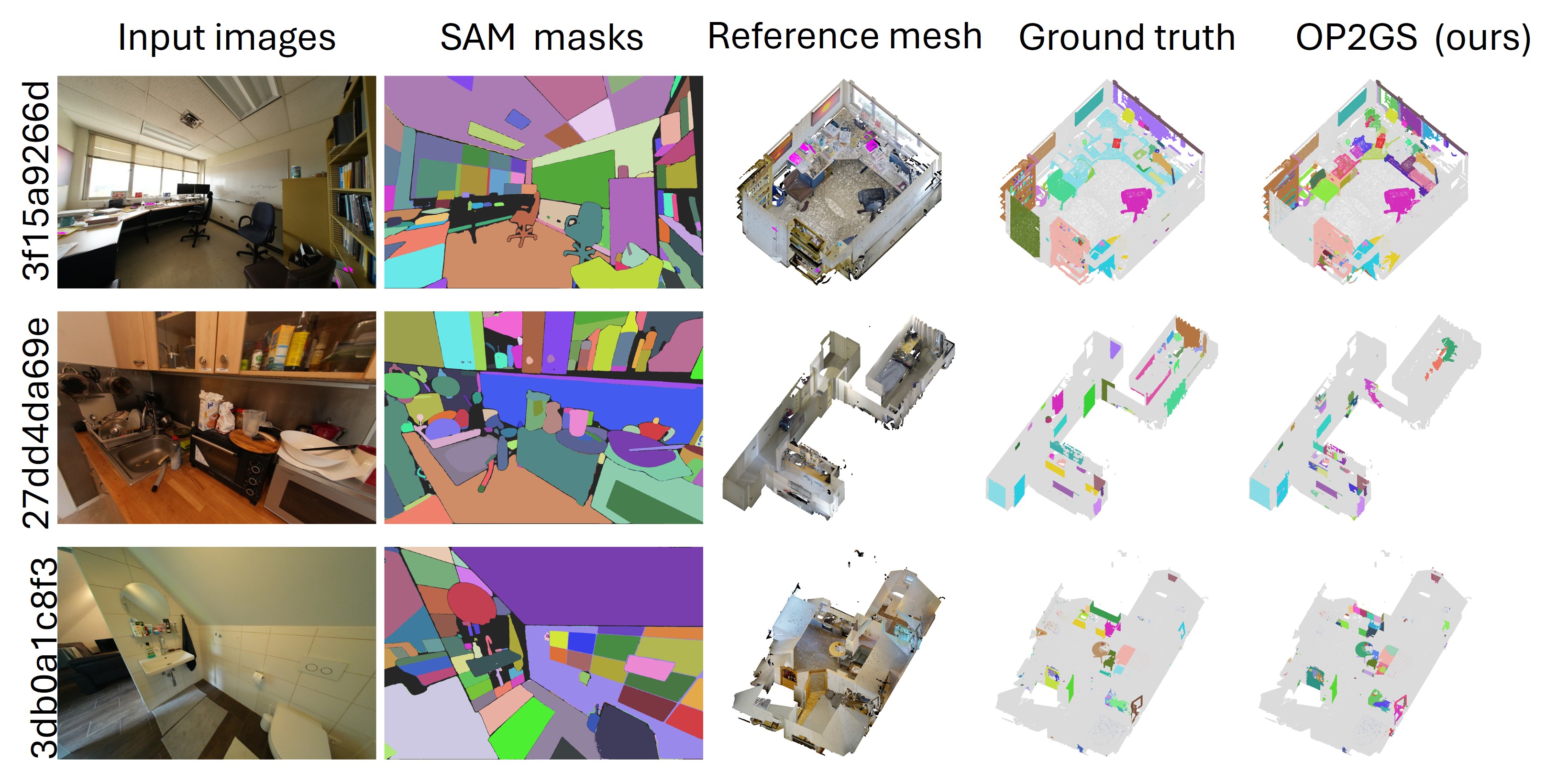}
  \caption{\textbf{The three ScanNet++ scenes with the lowest AP} (SAM masks): 3f15a9266d (4.6 AP), 27dd4da69e (5.8 AP) and 3db0a1c8f3 (6.3 AP). Ground-truth objects and the OP2GS predictions matched to them (IoU ${\ge}\,0.5$) share a color; unlabeled regions and predictions lying mostly on them are gray. The low AP comes from the input masks: SAM over-segments these scenes, splitting objects and large surfaces into many masks, so many predicted instances cannot be matched to a ground-truth object.}
  \label{fig:supp_hard}
\end{figure*}

\begin{figure*}[t]
  \centering
  \includegraphics[width=0.9\linewidth]{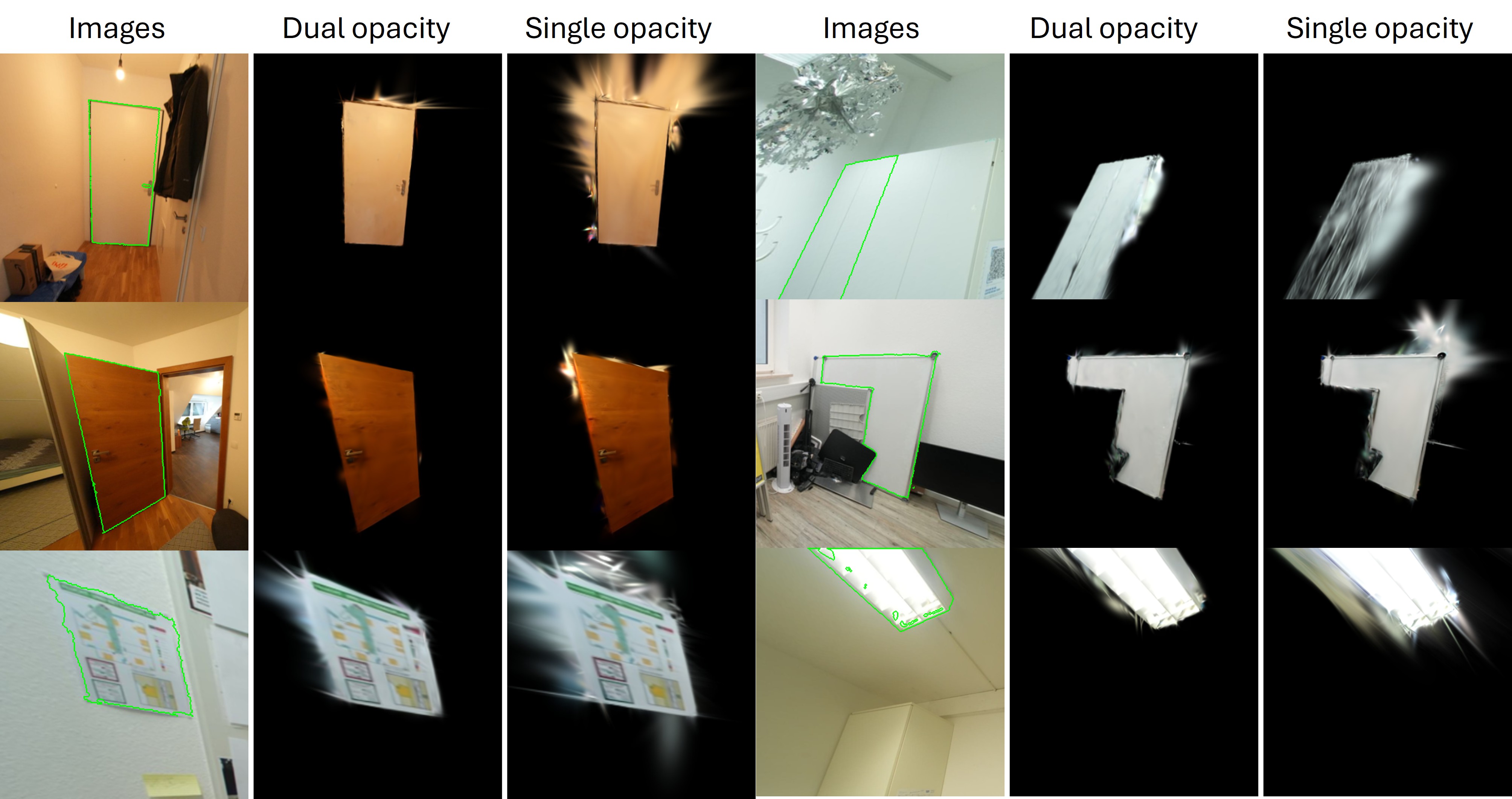}
  \caption{\textbf{Objects extracted with dual vs.\ single opacity} on ScanNet++ (SAM masks, same pipeline). Each object is rendered from the Gaussians of its best-matching label whose opacity exceeds 0.1 ($\sigma^{*}$ for dual, $\sigma$ for single opacity); green: ground-truth outline. With a single opacity, boundary Gaussians and floaters remain attached to the object as halos.}
  \label{fig:supp_dualsingle}
  \vspace{4mm}
  \includegraphics[width=0.9\linewidth]{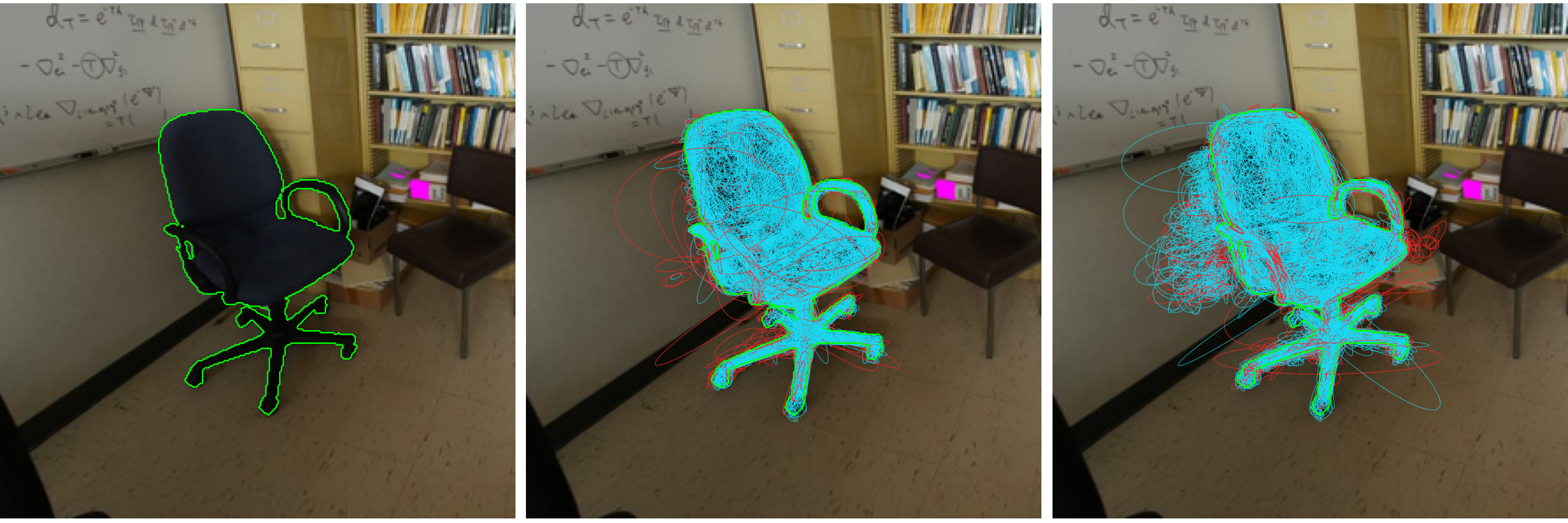}
  \caption{\textbf{Gaussians of one extracted object} (ScanNet++, a chair). Left: training view with the ground-truth outline (green). Middle: dual opacity; right: single opacity. Each Gaussian of the label that best matches the chair is drawn as its projected 2$\sigma$ ellipse: cyan if it is kept for extraction (opacity ${\ge}\,0.1$; $\sigma^{*}$ for dual, $\sigma$ for single opacity), red if it is removed. With dual opacity, the large Gaussians that reach beyond the chair onto the wall and the floor are removed; with a single opacity, many of them keep an opacity above the threshold and remain, which causes the halos in Fig.~\ref{fig:supp_dualsingle}.}
  \label{fig:supp_rings}
\end{figure*}

\cref{fig:vis_scannetpp} compares full scenes of OP2GS, OpenSplat3D$^\dagger$ and Gaussian Grouping; Gaussian Grouping is shown only qualitatively, run with its official SAM and DEVA pipeline on the same frames.
\cref{fig:supp_hard} shows the three scenes with the lowest AP (\cref{tab:supp_perscene_spp}).
\cref{fig:supp_dualsingle} compares objects extracted with dual and single opacity, and \cref{fig:supp_rings} shows the Gaussians behind one of them.

\begin{figure*}[t]
  \centering
  \includegraphics[width=0.82\linewidth]{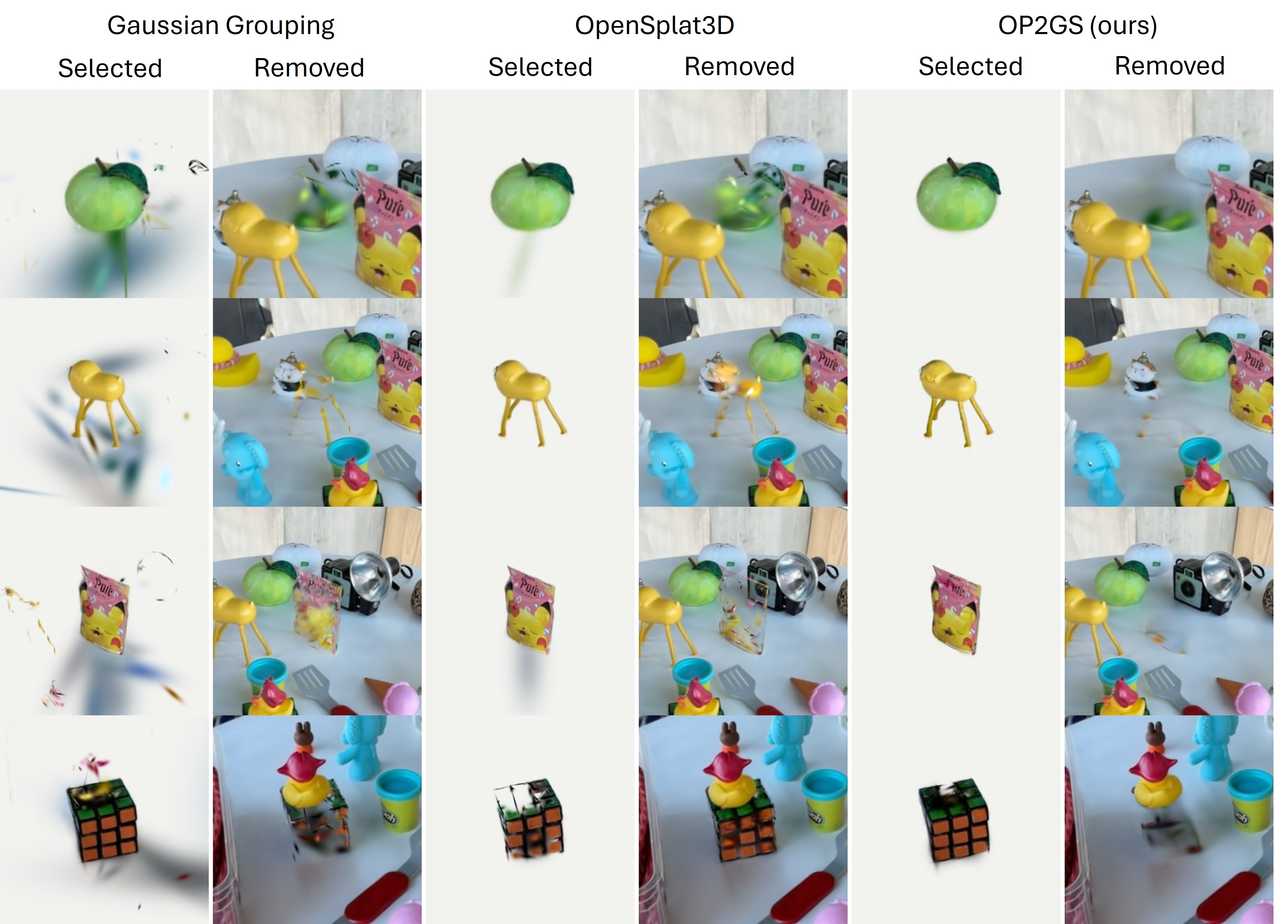}
  \caption{\textbf{Object removal on LERF figurines.} Selected: only the Gaussians selected for the object. Removed: the scene after these Gaussians are deleted. Gaussian Grouping selects floaters far from the object and leaves part of it; OpenSplat3D leaves a ghost of the object; OP2GS removes the object almost completely.}
  \label{fig:supp_removal}
\end{figure*}

\paragraph{Object removal.}
\cref{fig:supp_removal} deletes one object on LERF figurines with two feature-based methods: Gaussian Grouping~\cite{ye2023gaussiangrouping} (identity features from DEVA masks) and OpenSplat3D~\cite{piekenbrinck2025opensplat3d} (contrastive features on SAM masks).
Baseline features are learned by gradient descent, and the gradient reaching a Gaussian scales with its rendering weight $\alpha_i T_i$.
Gaussians inside the object have small weights (\eg, $10^{-3}$), can keep features close to their initialization, and remain as remnants after deletion.
OP2GS lifts its labels by the voting rule of \cref{sec:assoc}, $l_i = \arg\max_k S_{ik}/A_{ik}$, where both terms are sums of the Gaussian's own weights; a small weight scales numerator and denominator alike and cancels in the ratio.
A Gaussian inside the apple therefore still takes the apple's label, and Gaussians that are never visible take the label of their nearest neighbor.
Deleting a label removes the interior together with the surface.
Concurrent work, VODA~\cite{cho2026voda}, identifies the same under-supervision in feature-based methods and adds a volumetric loss; OP2GS avoids it by construction.

\begin{figure*}[t]
  \centering
  \includegraphics[width=\linewidth]{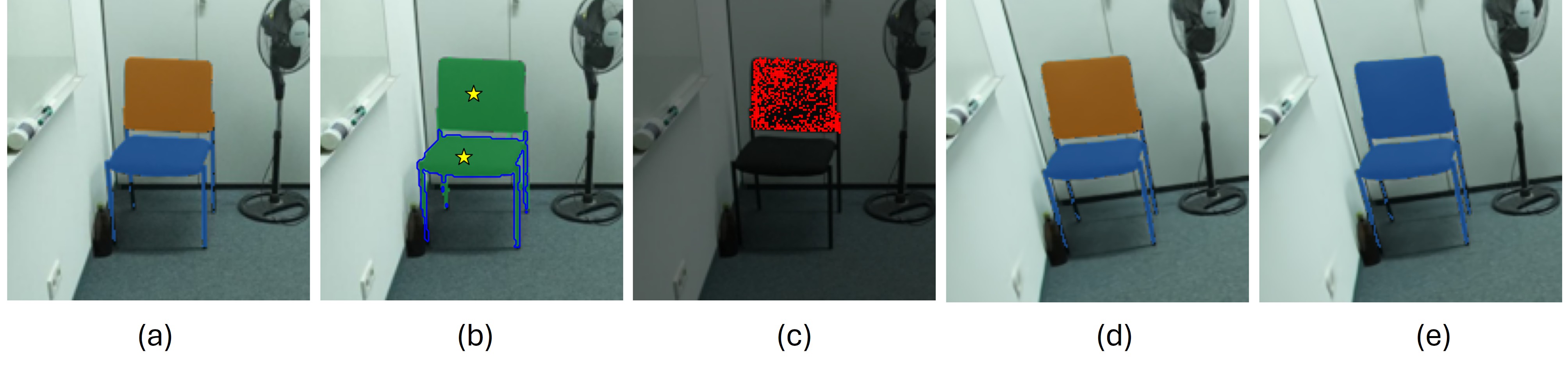}
  \caption{\textbf{Application: correcting an instance with two clicks} (ScanNet++). (a)~Current result: the chair (blue) misses its backrest (orange). (b)~Two clicks prompt SAM for a mask of the whole chair (green) in this view. (c)~The Gaussians that render into the mask (red) are relabeled to the chair; no retraining is needed, because labels are discrete. (d)~A novel view before and (e)~after the correction.}
  \label{fig:supp_click}
\end{figure*}

\paragraph{Interactive correction.}
Because every Gaussian carries a discrete label, a missed part of an object can be corrected directly: two clicks prompt SAM for the object mask in one view, and the Gaussians that render into this mask are relabeled (\cref{fig:supp_click}).

\section{Convergence Protocol}
\label{sec:supp_frozen}
\label{sec:supp_convergence}
In the frozen convergence experiment (Fig.~\ref{fig:convergence}), the six scenes are randomly chosen from the 50 validation scenes. Every checkpoint is exported to the mesh with the same nearest-Gaussian transfer; each method keeps only Gaussians that its own opacity marks as present ($\sigma^{*}{>}0.1$ for OP2GS, $\alpha{>}0.1$ for Gaussian Grouping, HDBSCAN non-noise points for OpenSplat3D). Without this filter, nearly transparent Gaussians dominate the output of Gaussian Grouping: their identity encodings receive little gradient but still carry object labels.
Per-step time on one H200 GPU: 20\,ms (OP2GS) vs.\ 17\,ms (OpenSplat3D).

\clearpage


\begin{table}[t]
\centering
\caption{\textbf{Per-scene results of OP2GS on ScanNet++} (validation set, class-agnostic, \%), same models and readout ($\sigma^{*}{>}0.1$) as Tab.~\ref{tab:scannetpp}. Each row is evaluated on its scene alone with the official evaluation; the last row pools all 50 scenes as in the main paper, which is not the mean of the rows. EntitySeg masks improve 39 of the 50 scenes and lower 10, so the +2.9 AP of Tab.~\ref{tab:scannetpp} is not driven by a few scenes.}
\label{tab:supp_perscene_spp}
\setlength{\tabcolsep}{2pt}
\scriptsize
\begin{tabular*}{\linewidth}{@{\extracolsep{\fill}}lcccccc@{}}
\toprule
 & \multicolumn{3}{c}{SAM masks} & \multicolumn{3}{c}{EntitySeg masks} \\
\cmidrule(lr){2-4}\cmidrule(l){5-7}
Scene & AP & AP$_{50}$ & AP$_{25}$ & AP & AP$_{50}$ & AP$_{25}$ \\
\midrule
\texttt{09c1414f1b} & 11.8 & 25.4 & 40.6 & 11.3 & 24.8 & 43.7 \\
\texttt{0d2ee665be} & 18.7 & 34.0 & 62.1 & 20.4 & 43.6 & 70.0 \\
\texttt{13c3e046d7} & 31.0 & 50.0 & 51.6 & 30.2 & 47.7 & 68.2 \\
\texttt{1ada7a0617} & 36.4 & 57.9 & 71.7 & 41.1 & 59.5 & 69.8 \\
\texttt{21d970d8de} & 34.2 & 57.0 & 70.8 & 48.1 & 64.9 & 77.1 \\
\texttt{25f3b7a318} & 18.1 & 33.9 & 54.6 & 23.3 & 36.0 & 61.1 \\
\texttt{27dd4da69e} & 5.8 & 14.1 & 28.3 & 5.9 & 12.8 & 33.6 \\
\texttt{286b55a2bf} & 24.7 & 49.1 & 55.1 & 30.2 & 55.1 & 70.5 \\
\texttt{31a2c91c43} & 25.1 & 41.2 & 73.8 & 25.3 & 52.2 & 78.6 \\
\texttt{3864514494} & 14.5 & 29.4 & 52.2 & 13.4 & 29.0 & 50.7 \\
\texttt{38d58a7a31} & 35.1 & 62.9 & 78.5 & 39.1 & 62.0 & 73.7 \\
\texttt{3db0a1c8f3} & 6.3 & 14.0 & 39.8 & 9.6 & 25.1 & 51.6 \\
\texttt{3e8bba0176} & 29.0 & 51.5 & 67.6 & 31.3 & 55.2 & 64.4 \\
\texttt{3f15a9266d} & 4.6 & 7.6 & 11.5 & 4.9 & 8.7 & 20.5 \\
\texttt{40aec5fffa} & 29.3 & 44.3 & 70.0 & 19.9 & 27.6 & 55.9 \\
\texttt{45b0dac5e3} & 26.8 & 49.4 & 78.8 & 27.9 & 48.4 & 73.5 \\
\texttt{5748ce6f01} & 20.4 & 36.2 & 67.2 & 27.2 & 50.0 & 66.7 \\
\texttt{578511c8a9} & 17.8 & 41.1 & 71.1 & 18.8 & 39.8 & 61.4 \\
\texttt{5942004064} & 22.9 & 46.1 & 70.2 & 22.4 & 44.6 & 59.1 \\
\texttt{5eb31827b7} & 21.4 & 39.6 & 51.5 & 22.8 & 39.8 & 57.5 \\
\texttt{5ee7c22ba0} & 41.6 & 69.1 & 80.0 & 24.8 & 47.5 & 63.2 \\
\texttt{5f99900f09} & 24.1 & 42.8 & 56.1 & 34.1 & 50.5 & 63.3 \\
\texttt{6115eddb86} & 17.9 & 39.0 & 58.4 & 27.8 & 49.7 & 63.7 \\
\texttt{7831862f02} & 37.6 & 48.2 & 60.3 & 51.2 & 73.6 & 76.0 \\
\texttt{7b6477cb95} & 27.0 & 42.7 & 61.0 & 33.7 & 52.4 & 58.5 \\
\texttt{7bc286c1b6} & 42.2 & 70.3 & 79.4 & 34.7 & 48.6 & 64.6 \\
\texttt{825d228aec} & 30.8 & 45.7 & 65.6 & 25.5 & 54.7 & 61.1 \\
\texttt{9071e139d9} & 14.6 & 28.6 & 38.9 & 25.3 & 46.8 & 52.6 \\
\texttt{99fa5c25e1} & 27.6 & 42.4 & 67.9 & 24.7 & 41.4 & 83.0 \\
\texttt{a24f64f7fb} & 31.4 & 46.0 & 55.9 & 37.0 & 55.8 & 60.4 \\
\texttt{a8bf42d646} & 16.6 & 28.6 & 52.8 & 23.1 & 41.9 & 49.8 \\
\texttt{a980334473} & 17.2 & 34.0 & 67.5 & 20.2 & 37.0 & 52.7 \\
\texttt{ac48a9b736} & 16.4 & 36.0 & 63.0 & 19.4 & 43.8 & 65.0 \\
\texttt{acd95847c5} & 28.9 & 49.9 & 63.1 & 29.6 & 49.5 & 61.9 \\
\texttt{b0a08200c9} & 26.9 & 49.3 & 72.9 & 31.8 & 53.1 & 75.8 \\
\texttt{bcd2436daf} & 28.2 & 48.0 & 60.6 & 29.3 & 42.9 & 57.4 \\
\texttt{bde1e479ad} & 37.6 & 64.2 & 80.4 & 49.0 & 70.4 & 77.3 \\
\texttt{c49a8c6cff} & 33.2 & 57.3 & 80.1 & 34.1 & 61.7 & 77.6 \\
\texttt{c4c04e6d6c} & 29.8 & 48.5 & 66.4 & 36.6 & 61.4 & 74.0 \\
\texttt{c50d2d1d42} & 25.9 & 45.0 & 66.2 & 33.6 & 52.8 & 71.5 \\
\texttt{c5439f4607} & 25.1 & 47.1 & 62.3 & 28.6 & 55.9 & 70.7 \\
\texttt{cc5237fd77} & 18.1 & 31.3 & 43.9 & 18.3 & 33.1 & 52.2 \\
\texttt{d755b3d9d8} & 20.0 & 36.8 & 56.7 & 23.5 & 40.3 & 57.2 \\
\texttt{e398684d27} & 21.6 & 46.7 & 76.2 & 22.1 & 45.4 & 64.0 \\
\texttt{e7af285f7d} & 13.1 & 27.0 & 38.6 & 12.9 & 32.2 & 51.8 \\
\texttt{f2dc06b1d2} & 20.4 & 38.3 & 54.8 & 20.4 & 44.5 & 61.8 \\
\texttt{f3685d06a9} & 19.7 & 43.0 & 56.5 & 21.4 & 47.1 & 58.3 \\
\texttt{f3d64c30f8} & 17.7 & 38.4 & 56.3 & 22.3 & 44.2 & 64.1 \\
\texttt{f9f95681fd} & 31.2 & 46.4 & 67.4 & 34.6 & 47.8 & 63.6 \\
\texttt{fb5a96b1a2} & 25.1 & 53.2 & 64.1 & 27.2 & 56.3 & 65.4 \\
\midrule
All scenes pooled & 20.2 & 37.1 & 54.7 & 23.1 & 40.9 & 56.8 \\
\bottomrule
\end{tabular*}
\end{table}

\begin{table}[t]
\centering
\caption{\textbf{Per-scene results of OP2GS on Replica} (Gaga protocol, \%), sum readout as in Tab.~\ref{tab:replica}; the last row is the mean over the eight scenes reported in the main paper.}
\label{tab:supp_perscene_replica}
\setlength{\tabcolsep}{2pt}
\footnotesize
\begin{tabular*}{\linewidth}{@{\extracolsep{\fill}}lcccccc@{}}
\toprule
 & \multicolumn{3}{c}{SAM masks} & \multicolumn{3}{c}{EntitySeg masks} \\
\cmidrule(lr){2-4}\cmidrule(l){5-7}
Scene & mIoU & Prec. & Recall & mIoU & Prec. & Recall \\
\midrule
office\_0 & 44.34 & 22.73 & 50.00 & 38.02 & 30.95 & 43.33 \\
office\_1 & 52.99 & 31.71 & 59.09 & 48.88 & 36.23 & 56.82 \\
office\_2 & 53.96 & 28.77 & 60.87 & 55.02 & 39.45 & 62.32 \\
office\_3 & 59.32 & 35.90 & 68.29 & 53.00 & 36.64 & 58.54 \\
office\_4 & 53.89 & 43.04 & 60.71 & 45.15 & 38.67 & 51.79 \\
room\_0 & 50.06 & 38.76 & 53.76 & 46.65 & 39.67 & 51.61 \\
room\_1 & 64.54 & 53.25 & 75.93 & 53.11 & 40.74 & 61.11 \\
room\_2 & 55.55 & 44.19 & 61.29 & 52.89 & 40.91 & 58.06 \\
\midrule
Mean & 54.33 & 37.29 & 61.24 & 49.09 & 37.91 & 55.45 \\
\bottomrule
\end{tabular*}
\end{table}

\begin{table}[t]
\centering
\caption{\textbf{Per-scene results of OP2GS on ScanNet} (10 scenes, \%), $\sigma^{*}{>}0.1$ readout without post-processing as in Tab.~\ref{tab:scannet}; the last row is the mean over the ten scenes reported in the main paper.}
\label{tab:supp_perscene_scannet}
\setlength{\tabcolsep}{1pt}
\scriptsize
\begin{tabular*}{\linewidth}{@{\extracolsep{\fill}}lcccccc@{}}
\toprule
 & \multicolumn{3}{c}{SAM masks} & \multicolumn{3}{c}{EntitySeg masks} \\
\cmidrule(lr){2-4}\cmidrule(l){5-7}
Scene & mIoU & mAcc@.5 & mAcc@.25 & mIoU & mAcc@.5 & mAcc@.25 \\
\midrule
scene0000\_00 & 42.66 & 44.78 & 74.63 & 42.89 & 40.30 & 71.64 \\
scene0062\_00 & 54.78 & 54.17 & 91.67 & 49.31 & 54.17 & 83.33 \\
scene0070\_00 & 54.29 & 43.48 & 91.30 & 47.95 & 30.43 & 78.26 \\
scene0097\_00 & 57.20 & 65.22 & 100.00 & 55.84 & 60.87 & 91.30 \\
scene0140\_00 & 43.33 & 21.28 & 89.36 & 61.07 & 72.34 & 89.36 \\
scene0200\_00 & 57.16 & 68.42 & 89.47 & 47.03 & 42.11 & 78.95 \\
scene0347\_00 & 53.93 & 53.57 & 85.71 & 58.49 & 67.86 & 92.86 \\
scene0400\_00 & 50.48 & 34.78 & 82.61 & 51.52 & 47.83 & 82.61 \\
scene0590\_00 & 54.62 & 60.00 & 91.67 & 53.37 & 58.33 & 95.00 \\
scene0645\_00 & 50.54 & 50.00 & 82.56 & 47.77 & 45.35 & 74.42 \\
\midrule
Mean & 51.90 & 49.57 & 87.90 & 51.52 & 51.96 & 83.77 \\
\bottomrule
\end{tabular*}
\end{table}

\clearpage 
{
    \small
    \bibliographystyle{ieeenat_fullname}
    \bibliography{main}
}

\end{document}